\documentclass[sigconf]{acmart}
\AtBeginDocument{%
  }

\setcopyright{acmlicensed}
\copyrightyear{2018}
\acmYear{2018}
\acmDOI{XXXXXXX.XXXXXXX}
\acmConference[Conference acronym 'XX]{Make sure to enter the correct
  conference title from your rights confirmation email}{June 03--05,
  2018}{Woodstock, NY}
\acmISBN{978-1-4503-XXXX-X/2018/06}

\usepackage{enumitem}
\usepackage{threeparttable}
\usepackage{pifont}
\usepackage{multirow}
\usepackage{makecell}

\usepackage{tabularx}

\usepackage{booktabs}
\usepackage{array}
\usepackage{pifont}
\usepackage{xcolor}

\newcommand{\cmark}{\textcolor{green!50!black}{\ding{51}}}
\newcommand{\xmark}{\textcolor{red!65!black}{\ding{55}}}
\newcommand{\pmark}{\textcolor{orange!80!black}{\ding{108}}}

\usepackage{adjustbox}

\newcommand{\apptablesingle}{%
  \small
  \setlength{\tabcolsep}{4pt}%
  \renewcommand{\arraystretch}{1.08}%
}

\newcommand{\apptablewide}{%
  \scriptsize
  \setlength{\tabcolsep}{2.2pt}%
  \renewcommand{\arraystretch}{1.05}%
}

\usepackage{tcolorbox}
\tcbset{colback=white, colframe=black, boxrule=0.6pt, arc=1pt, left=6pt, right=6pt, top=6pt, bottom=6pt}

\definecolor{lightpurple}{RGB}{230, 224, 244}
\definecolor{lightorange}{RGB}{255, 236, 214}

\begin{document}

\title[TimeLitmus]{
TimeLitmus: A Diagnostic Benchmark for Cross-Modal Understanding and Explanation Faithfulness in Event-Conditioned Time-Series Prediction}


\author{Jie Gong}
\affiliation{%
  \institution{School of Artificial Intelligence, Wuhan University}
  \country{China}
}

\author{Maowei Jiang}
\affiliation{%
  \institution{Nanjing Audit University}
  \country{China}
}

\author{Zhiwei Liu}
\affiliation{%
  \institution{The University of Manchester}
  \country{United Kingdom}
}

\author{Yankai Chen}

\affiliation{%
  \institution{MBZUAI}
  \country{United Arab Emirates}
}

\affiliation{%
  \institution{McGill University}
  \country{Canada}
}

\author{Guojun Xiong}
\affiliation{%
  \institution{School of Computer Science, Shanghai Jiao Tong University}
  \country{China}
}

\author{Xue Liu}
\affiliation{%
  \institution{MBZUAI}
  \country{United Arab Emirates}
}

\affiliation{%
  \institution{McGill University}
  \country{Canada}
}

\author{Min Peng}
\affiliation{%
  \institution{School of Artificial Intelligence, Wuhan University}
  \country{China}
}

\author{Qianqian Xie}
\affiliation{%
  \institution{School of Artificial Intelligence, Wuhan University}
  \country{China}
}
\email{xieq@whu.edu.cn}
\authornote{Corresponding author.}

\author{Sophia Ananiadou}
\affiliation{%
  \institution{The University of Manchester}
  \country{United Kingdom}
}

\renewcommand{\shortauthors}{Jie Gong et al.}

\begin{abstract}

Large language models (LLMs) are increasingly used to make predictions
from numerical time-series histories and textual events. Yet accuracy alone
cannot reveal whether correct answers reflect effective integration of the
two inputs or instead arise from event polarity, unimodal priors, or
superficial cues. Likewise, plausible explanations may rationalize
predictions without faithfully reflecting the evidence that drives model
behavior. We introduce TimeLitmus, a diagnostic benchmark for cross-modal
understanding and explanation faithfulness in event-conditioned time-series
prediction. TimeLitmus contains 4,856 evaluation records across Finance and
Traffic, combining natural prediction with controlled counterfactual and
contrastive interventions, explanation-targeted faithfulness tests, and
systematic shortcut controls.
Across ten representative LLMs, standard prediction accuracy substantially
overstates reliable cross-modal understanding: Hard Paired Contrast (HPC)
pair correctness peaks at only 19.2\% in Finance and 11.7\% in Traffic,
and all ten models show lower-than-expected consistency on Finance
series-side controls. Models often recognize scenario relations explicitly
yet fail to apply them during independent prediction. Explanation
faithfulness shows a similar gap: in Traffic, most models cite the
manipulated temporal factor in over 90\% of cases, while behavioral support
remains below 22\%. Human annotators outperform LLMs on matched controlled
and hard-pair diagnostics, confirming that these distinctions are
recoverable from the inputs. Natural-only adaptation yields selective gains
in evidence selection and input sensitivity, but not consistent gains in
controlled or hard-pair behavior. The benchmark, evaluation suite, and supervised adaptation data
will be released publicly.
\end{abstract}




\maketitle

\section{Introduction}

\begin{figure}[t]
    \centering
    \includegraphics[
        width=1\linewidth,
        height=0.36\textheight,
        keepaspectratio
    ]{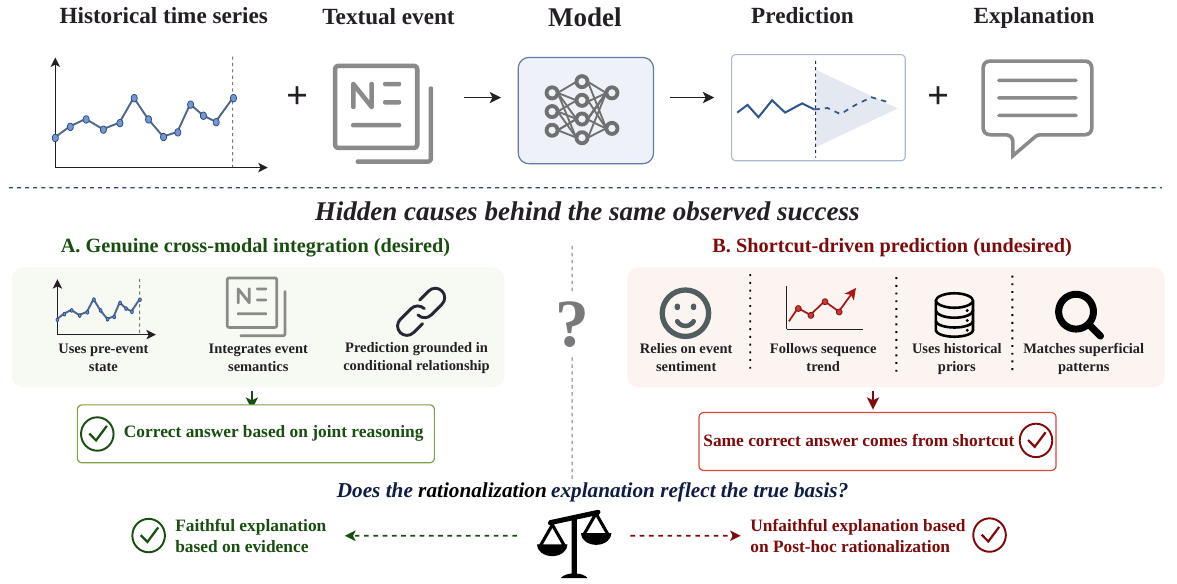}
    \caption{Why accuracy alone is insufficient. Correct predictions and plausible explanations may result from either genuine cross-modal integration or shortcut cues.}
    \label{fig:1}
\end{figure}

When large language models make correct predictions from historical time series and textual events, the central evaluation question is whether their accuracy reflects genuine cross-modal integration or reliance on unimodal cues, event sentiment, historical priors, or shallow statistical matching \citep{yuan2024llms}. 
A related question is whether a plausible explanation truly reflects the basis of prediction, or merely rationalizes an answer whose actual basis lies elsewhere \citep{turpin2023language}. In this setting, event-conditioned time-series prediction refers to predicting
a future response jointly from a pre-event time-series history and a textual event. Time-series data are semantically underspecified: they record how variables evolve, but rarely reveal the external events, generative processes, or contexts behind these changes.
Textual events can provide such missing semantic conditions \citep{xu2025beyond,williams2024context}. Yet prediction depends on the conditional relationship between the historical trajectory and the textual event: the same trajectory may imply different futures under different events, and the same event may have different implications under different time-series states. As a result, evaluations that rely only on prediction correctness or explanation plausibility may overestimate a model’s cross-modal understanding of time series and text. In high-stakes domains such as finance, healthcare, and public safety, models that produce correct answers for the wrong reasons may pass standard evaluations but fail when conditions change. Plausible yet unfaithful explanations can further create false confidence and mislead downstream decisions, potentially leading to substantial financial losses, harmful clinical decisions, or serious public-safety consequences.

Existing benchmarks on time series-text modeling and time-series reasoning have broadened the evaluation of models over numerical sequences, textual contexts, and reasoning tasks.
However, they do not directly determine whether correct answers come from the intended conditional relationship between historical time series and textual events, or whether claims in generated explanations receive corresponding behavioral support. On the prediction side, multimodal and contextual time-series benchmarks such as MTBench and TemporalBench evaluate forecasting, trend analysis, news-driven QA, or event-informed reasoning under textual contexts \citep{williams2024context,liu2024time,chen2025mtbench,weng2026temporalbench}.
Time-series QA and reasoning benchmarks such as Time-MQA and TSRBench cast time-series tasks into natural-language QA or general reasoning formats \citep{kong2025time,gwiazda2026timeseriesexamagent,yu2026tsrbench}. These benchmarks primarily measure task performance or answer correctness, rather than diagnosing whether correct outputs arise from genuine cross-modal integration or from unimodal cues and shallow statistical matching. 
For explanations, reasoning-trace benchmarks such as TFRBench evaluate generated reasoning processes for forecasting systems \citep{ahamed2026tfrbench,sivalingam2026llm}. 
They mainly assess whether traces are complete, coherent, or superficially aligned with the answer, leaving open whether the cited factors are supported by the model's prediction behavior.

\begin{table}[t]
\centering

\caption{
Comparison with representative evaluation frameworks.
\textbf{Event--Series}: joint temporal--text conditioning;
\textbf{Controlled}: targeted interventions with predefined effects;
\textbf{Hard Pairs}: natural pairs requiring distinct responses;
\textbf{Shortcut}: invariance and partial-input tests;
\textbf{Claim-Linked}: behavioral support for cited factors.
\cmark{}, \xmark{}, and \pmark{} denote explicit, absent, and partial
coverage, respectively.
}

\label{tab:related-comparison}

\scriptsize
\setlength{\tabcolsep}{2.7pt}
\renewcommand{\arraystretch}{1.15}

\resizebox{\columnwidth}{!}{%
\begin{tabular}{@{}lccccc@{}}
\toprule

\textbf{Work}
& \shortstack{\textbf{Event--}\\\textbf{Series}}
& \shortstack{\textbf{Controlled}\\\textbf{Pairs}}
& \shortstack{\textbf{Hard}\\\textbf{Pairs}}
& \shortstack{\textbf{Shortcut}\\\textbf{Tests}}
& \shortstack{\textbf{Claim-Linked}\\\textbf{Faithfulness}}
\\

\midrule

\multicolumn{6}{@{}l}{
\textit{Event-conditioned time-series prediction and reasoning}
}
\\[1pt]

Time-MMD~\citep{liu2024time}
& \cmark
& \xmark
& \xmark
& \xmark
& \xmark
\\

Context is Key~\citep{williams2024context}
& \cmark
& \xmark
& \xmark
& \pmark
& \xmark
\\

TemporalBench~\citep{weng2026temporalbench}
& \cmark
& \pmark
& \xmark
& \pmark
& \xmark
\\

\midrule

\multicolumn{6}{@{}l}{
\textit{Behavioral diagnostics beyond aggregate accuracy}
}
\\[1pt]

Contrast Sets~\citep{gardner2020evaluating}
& \xmark
& \cmark
& \xmark
& \pmark
& \xmark
\\

CheckList~\citep{ribeiro2020beyond}
& \xmark
& \pmark
& \xmark
& \cmark
& \xmark
\\

\midrule

\multicolumn{6}{@{}l}{
\textit{Explanation and reasoning evaluation}
}
\\[1pt]

ERASER~\citep{deyoung2020eraser}
& \xmark
& \xmark
& \xmark
& \xmark
& \pmark
\\

XForecast~\citep{aksu2024xforecast}
& \xmark
& \xmark
& \xmark
& \xmark
& \pmark
\\

TFRBench~\citep{ahamed2026tfrbench}
& \pmark
& \xmark
& \xmark
& \xmark
& \pmark
\\

\midrule

\textbf{TimeLitmus}
& \cmark
& \cmark
& \cmark
& \cmark
& \cmark
\\

\bottomrule
\end{tabular}%
}

\end{table}

To address these evaluation gaps, we introduce TimeLitmus, a diagnostic benchmark for cross-modal understanding and explanation faithfulness in event-conditioned time-series prediction. 
As summarized in
Table~\ref{tab:related-comparison}, TimeLitmus brings these complementary
diagnostic perspectives into a unified benchmark.
TimeLitmus contains 4,856 evaluation records across Finance and Traffic. Each instance combines a historical numerical sequence, a textual event, and a prediction question, requiring both a future response judgment and a natural-language explanation. 
Because aggregate accuracy cannot reveal whether a correct prediction depends on the intended relationship between the two modalities, TimeLitmus constructs paired-contrastive and counterfactual variants.
These variants selectively modify event presence, direction, or strength and the pre-event time-series state while preserving the surrounding task structure. Prediction changes consistent with these controlled interventions provide behavioral evidence that the model uses the conditional event–series relationship \citep{gardner2020evaluating}. Inconsistent behavior may instead reveal shortcut-driven correctness, such as relying only on event polarity, extrapolating the historical trend while ignoring the event, or exploiting common outcome priors and superficial textual cues. Because a plausible explanation may still rationalize a prediction made on other grounds, TimeLitmus further intervenes on the event or temporal factors cited in the explanation and tests whether the prediction responds accordingly \citep{turpin2023language}. 
The benchmark also introduces unimodal, anonymized, polarity-based, and hint-control settings to isolate shortcut reliance.
It evaluates prediction correctness, counterfactual mechanism consistency, stated-rationale faithfulness, and shortcut robustness separately. We additionally construct a separate natural-prediction supervision corpus and use it to study whether natural-only QLoRA adaptation transfers to the capabilities diagnosed by TimeLitmus. For open-weight models, a masking-based output-logit sensitivity analysis further examines whether adaptation changes how individual input components support prediction confidence.

Using TimeLitmus, we evaluate representative LLMs and find that standard accuracy can substantially overestimate their cross-modal understanding. Models that perform well on natural instances often become inconsistent under counterfactual or shortcut-control settings, suggesting reliance on unimodal cues or surface-level matching rather than effective use of the event–series relationship. They may also identify relevant evidence or distinguish scenario relations when explicitly prompted, yet fail to apply the same distinctions during independent prediction. Plausible explanations frequently lack corresponding behavioral support, showing that explanation plausibility alone is insufficient evidence of faithfulness. Natural-only QLoRA adaptation yields selective gains, primarily in evidence selection, but does not consistently improve controlled consistency or hard-pair behavior. Output-logit probing further shows that adaptation can increase confidence-level sensitivity to intended event, series, and evidence inputs without reliably translating these shifts into relation-consistent predictions.

Our contributions are as follows:
\textbf{(i) Correctness attribution and explanation faithfulness.}
We formulate two diagnostic evaluation problems for LLMs in event-conditioned time-series prediction: whether correct predictions arise from effective use of the conditional relationship between historical time series and textual events, and whether the factors cited in generated explanations receive corresponding behavioral support.
\textbf{(ii) TimeLitmus benchmark and adaptation resources.}
We introduce a 4,856-record benchmark across Finance and Traffic,
combining controlled counterfactuals, hard contrasts,
explanation-targeted interventions, and shortcut controls to evaluate
prediction grounding and explanation faithfulness. We also construct
a separate natural-prediction supervision corpus for studying adaptation
transfer to the capabilities diagnosed by TimeLitmus.
\textbf{(iii) Systematic evaluation and adaptation findings.} We systematically evaluate representative LLMs and study whether natural-only QLoRA adaptation transfers to the capabilities diagnosed by TimeLitmus. Behavioral evaluation and output-logit sensitivity analysis reveal separations among natural accuracy, evidence recognition, confidence-level input sensitivity, relation-consistent prediction, and explanation faithfulness, showing that adaptation produces selective gains without consistent transfer to controlled or hard-pair behavior.

\section{Related Work}

\textbf{Event-Conditioned Time-Series Prediction and Evaluation.}
Recent work has extended time-series evaluation beyond numerical forecasting
to feature understanding, question answering, and reasoning with textual
context \citep{fons2024evaluating,kong2025time,yu2026tsrbench,sen2025bedtime,xie2023pixiu,xie2024finben}.
Time-MMD and the TimeText Corpus align numerical sequences with textual
information for multimodal time-series modeling
\citep{liu2024time,kim2024multi,xu2025beyond,huang2024open,xie2023wall}. Benchmarks including Context is Key,
MTBench, and TemporalBench evaluate forecasting or reasoning conditioned on
external descriptions and events
\citep{williams2024context,chen2025mtbench,weng2026temporalbench,wang2024news,jang2026if}.
Fidel-TS further emphasizes data integrity, leakage control, and benchmark
validity in multimodal forecasting \citep{xu2025fidel,liu2026rethinking}. These studies
primarily assess whether a model produces the correct answer or forecast.
They do not systematically distinguish genuine event--series integration from
success driven by either modality alone, dominant priors, or superficial cues.

\textbf{Diagnostic Evaluation beyond Accuracy.}
Research on shortcut learning and multimodal bias shows that strong benchmark
performance can arise from annotation artifacts or dominant unimodal signals
rather than the intended cross-modal relationship
\citep{geirhos2020shortcut,goyal2017making,zheng2025mllms}. Contrast sets,
counterfactually augmented data, and behavioral testing address this problem
by modifying task-relevant factors and examining whether predictions change
accordingly
\citep{gardner2020evaluating,kaushik2019learning,ribeiro2020beyond,niu2021counterfactual}.
TimeLitmus brings this diagnostic perspective to event-conditioned
time-series prediction, where the expected response depends jointly on event
semantics and the pre-event temporal state. It combines independently
evaluated controlled endpoints, natural hard paired contrasts, invariance
tests, and partial-modality controls to separate cross-modal integration from
shortcut success.

\textbf{Behaviorally Supported Explanations.}
Prior work distinguishes plausible explanations from faithful explanations
that reflect the factors supporting model behavior
\citep{jacovi2020towards,lyu2024towards,turpin2023language,lanham2023measuring,atanasova2023faithfulness}. General
explanation benchmarks such as ERASER evaluate evidence sufficiency and
comprehensiveness, while subsequent work studies behavioral consistency and
simulatability
\citep{deyoung2020eraser,parcalabescu2024measuring,hase2020evaluating,hase2020leakage}. 
In time-series settings,
XForecast evaluates explanation quality for forecasting, whereas TFRBench assesses
numerically grounded reasoning traces \citep{aksu2024xforecast,ahamed2026tfrbench}. 
Recent judge-based approaches
instead evaluate whether generated explanations correctly describe observed temporal
patterns \citep{sivalingam2026llm}.
TimeLitmus adds a claim-conditioned behavioral test: when an explanation cites
an event or temporal factor, controlled pairs test whether the model's
predictions follow the corresponding benchmark-defined transition. Together
with evidence validity and linked faithfulness, this separates correct factor
mention from behaviorally supported dependence.

%

\section{Problem Formulation}
\label{sec:problem}

TimeLitmus evaluates event-conditioned time-series prediction along four
complementary dimensions: natural-task correctness, consistency under
controlled changes, behavioral support for explanation claims, and shortcut
robustness. Appendix~\ref{app:notation} provides the complete indexed
definitions and notation.

\paragraph{Task.}
For domain $d\in\mathcal{D}$, let
$\mathbf{X}_d\in\mathbb{R}^{L\times p_d}$ denote an $L$-step pre-event
series with $p_d$ variables, $E_d\in\mathcal{E}_d$ a textual event,
$Q_d\in\mathcal{Q}_d$ the task presentation, and $\mathcal{Y}_d$ the
semantic response space. The benchmark labeling function
$g_d:\mathbb{R}^{L\times p_d}\times\mathcal{E}_d\rightarrow\mathcal{Y}_d$
assigns the expected response, while an evaluated model $M$ returns a
prediction and an explanation:
\begin{equation}
y_d=g_d(\mathbf{X}_d,E_d),
\qquad
(\hat{y}_d,O_d)=M(\mathbf{X}_d,E_d,Q_d).
\label{eq:base-task}
\end{equation}
Here, $y_d,\hat{y}_d\in\mathcal{Y}_d$ are the gold and predicted responses,
and $O_d\in\mathcal{O}_d$ is the generated explanation. Natural Accuracy is
the fraction of natural instances for which $\hat{y}_d=y_d$.

\paragraph{Controlled behavioral diagnostics.}
Let $m\in\{\mathrm{series},\mathrm{event}\}$ denote the intervention side.
For a fixed domain $d$ and side $m$, we suppress these two subscripts.
Let $N=N_{d,m}^{\mathrm{CF}}$ be the number of retained controlled pairs,
indexed by $i\in\{1,\ldots,N\}$. Each pair contains two independently
evaluated endpoints $b\in\{0,1\}$ with a shared task presentation $Q_i$.
A validated transformation $\tau_i$ changes only side $m$:
\begin{equation}
\left(\mathbf{X}_i^{(1)},E_i^{(1)}\right)
=
\tau_i\!\left(\mathbf{X}_i^{(0)},E_i^{(0)}\right),
\qquad
y_i^{(0)}\neq y_i^{(1)},
\label{eq:controlled-transformation}
\end{equation}
where
$y_i^{(b)}=g_d(\mathbf{X}_i^{(b)},E_i^{(b)})$.
The model is applied separately to each endpoint,
$(\hat{y}_i^{(b)},O_i^{(b)})
=M(\mathbf{X}_i^{(b)},E_i^{(b)},Q_i)$.

Let $\mathbb{I}[\cdot]$ be the indicator function. Pair $i$ is correct only
when both independently evaluated endpoints are correct:
\begin{equation}
B_i
=
\prod_{b\in\{0,1\}}
\mathbb{I}\!\left[\hat{y}_i^{(b)}=y_i^{(b)}\right],
\qquad
\mathrm{CF\text{-}PC}
=
\frac{1}{N}\sum_{i=1}^{N}B_i.
\label{eq:cf-pc}
\end{equation}
To distinguish pair-level inconsistency from endpoint difficulty, let
\begin{equation}
A^{(b)}
=
\frac{1}{N}\sum_{i=1}^{N}
\mathbb{I}\!\left[\hat{y}_i^{(b)}=y_i^{(b)}\right],
\qquad
\Delta_{\mathrm{CF}}
=
\mathrm{CF\text{-}PC}-A^{(0)}A^{(1)}.
\label{eq:delta-cf}
\end{equation}
Thus, $\Delta_{\mathrm{CF}}$ compares observed pair correctness with the rate
implied by the two endpoint accuracies under endpoint independence.

Label-preserving pairs retain the same gold response and are evaluated by
Prediction Invariance. Hard Paired Contrasts comprise naturally occurring,
similar endpoints with different gold responses and use the same
both-endpoints-correct criterion. Appendix~\ref{app:notation} gives their
complete definitions.

\paragraph{Claim-conditioned explanation faithfulness.}
Let $\mathcal{F}_d$ be the canonical evidence-factor vocabulary and
$\phi_d:\mathcal{O}_d\rightarrow2^{\mathcal{F}_d}$ map an explanation to
its cited factors. Within the same fixed $(d,m)$ stratum, let
$\varnothing\neq\mathcal{G}_i\subseteq\mathcal{F}_d$ be the factor-family
set manipulated in controlled pair $i$. The factors cited across the two
endpoint explanations and the corresponding citation indicator are
\begin{equation}
\mathcal{R}_i
=
\phi_d(O_i^{(0)})\cup\phi_d(O_i^{(1)}),
\qquad
C_i
=
\mathbb{I}\!\left[
\mathcal{R}_i\cap\mathcal{G}_i\neq\varnothing
\right].
\label{eq:claim-indicator}
\end{equation}
Citation Rate measures whether the manipulated factor is mentioned, while
Claim-Supported Dependence Rate measures whether such a claim is accompanied
by exact pair-level behavioral support:
\begin{equation}
\mathrm{Citation}
=
\frac{1}{N}\sum_{i=1}^{N}C_i,
\qquad
\mathrm{CSDR}
=
\frac{\sum_{i=1}^{N}C_iB_i}
     {\sum_{i=1}^{N}C_i}.
\label{eq:claim-metrics}
\end{equation}
CSDR is reported when $\sum_{i=1}^{N}C_i>0$. TimeLitmus complements this
claim-conditioned test with evidence-validity checks and linked-faithfulness
metrics that jointly require correct prediction and valid evidence
attribution.

\paragraph{Shortcut robustness.}
Shortcut robustness compares full-input prediction with event-only and
series-only inputs, and tests stability under label-preserving and
irrelevant-cue transformations. Together, these diagnostics separate
answer correctness, controlled cross-modal behavior, behaviorally supported
explanation claims, and shortcut reliance. Domain-specific response spaces
and transformations are introduced in Section~\ref{sec:benchmark};
complete metric and uncertainty definitions are provided in
Section~\ref{sec:metrics}, Appendix~\ref{app:metric-definitions}, and
Appendix~\ref{app:statistical-testing}.

\begin{figure}[t]
    \centering
    \includegraphics[
        width=0.98\linewidth,
        height=0.35\textheight,
        keepaspectratio
    ]{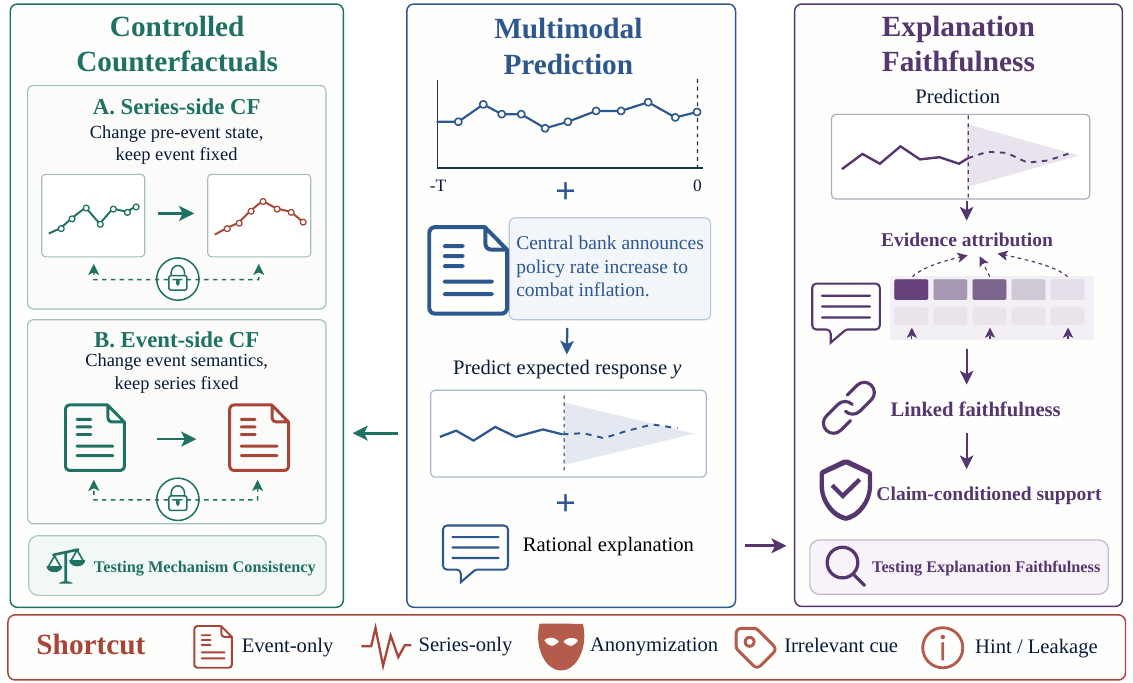}
    \caption{Overview of TimeLitmus. Controlled interventions and faithfulness diagnostics test whether event-conditioned predictions and explanations rely on the intended cross-modal evidence rather than shortcuts.}
    \label{fig:2}
\end{figure}

\section{TimeLitmus Benchmark}
\label{sec:benchmark}

\subsection{Overview}
\label{sec:benchmark-overview}

TimeLitmus instantiates the formulation in Section~\ref{sec:problem} in two
event-conditioned time-series domains: Finance and Traffic. Finance is built
on FNSPID~\citep{dong2024fnspid} and evaluates expected market responses to
financial events. Traffic is built on TraffiDent~\citep{gou2026traffident}
and evaluates expected traffic responses to roadway incidents. The two
domains share a unified diagnostic architecture while preserving
domain-specific response spaces, event semantics, temporal variables, and
rule predicates.

TimeLitmus evaluates the interaction between event semantics and the
pre-event temporal state. Most labels are rule-grounded expected responses,
which support auditable controlled interventions and strict expected-change
relations. Traffic natural-prediction and hard-paired-contrast labels are
instead derived from observed post-report response buckets. Realized
post-event observations are otherwise used as complementary validation
signals rather than as the primary definition of controlled labels.

The benchmark contains 4,856 task-specific evaluation records: 3,000 in
Finance and 1,856 in Traffic. These records are organized into four diagnostic
families: Natural Prediction, Correctness Attribution, Explanation
Faithfulness, and Shortcut Robustness. Pair- and link-based metrics are
computed over complete structured units rather than individual arm counts;
detailed record, endpoint, pair, and link accounting is provided in the
Appendix~\ref{app:expanded-evaluation-record-accounting}.
Across the two domains, TimeLitmus includes 800 natural-prediction records and
1,092 controlled expected-change records, corresponding to 546 strict pairs.
Among these pairs, 436 intervene on the pre-event series and 110 intervene on
event semantics. Both domains further include hard paired contrasts,
evidence-attribution records, linked-faithfulness records, shortcut controls,
and auxiliary relation probes.

\subsection{Diagnostic Design}
\label{sec:shared-diagnostic-design}

TimeLitmus separates natural-task performance from diagnostics that identify
the basis of model behavior. Natural Prediction evaluates full-input
expected-response prediction on observed event--series combinations.
Correctness Attribution tests whether successful predictions reflect
sensitivity to the relevant event semantics and temporal state.
Explanation Faithfulness evaluates whether cited evidence is valid
and behaviorally supported. Shortcut Robustness tests whether
performance depends on partial-modality inputs, irrelevant cues, or
response-preserving transformations.

\textbf{Controlled interventions.}
The central controlled diagnostics modify one modality while holding the
other fixed. Series-side interventions preserve the event while
changing the pre-event temporal state, testing whether the model adapts its
prediction to a different numerical context. Event-side
interventions preserve the time-series history while changing a
mechanism-relevant event attribute, testing whether the model responds to
event semantics under the same temporal context. Each retained
expected-change pair specifies a strict gold transition; pairs that could
legitimately support an unchanged prediction are excluded. The two endpoints
are presented and evaluated independently, so models cannot infer the pair
identity or intended transition from the prompt.

\textbf{Hard contrasts and relation recognition.}
Hard Paired Contrasts (HPC) compare realistic inputs that are
superficially similar but require different responses. They test whether a
model can discriminate difficult natural contrasts without access to explicit
intervention metadata. The Auxiliary Relation probe presents two
scenarios jointly and asks for their expected relation. It complements
independent endpoint evaluation by separating explicit contrast recognition
from consistent application during prediction.

\textbf{Invariance and shortcut controls.}
Invariance and irrelevant-cue diagnostics test whether predictions remain stable under changes that preserve the response or introduce non-mechanistic variation. Shortcut
diagnostics evaluate shallow solution paths such as series-only or event-only
prediction, dependence on irrelevant metadata, and sensitivity to
presentation changes. Together with controlled interventions, these controls
distinguish genuine event--series integration from performance attainable
through a single modality or superficial cue.

\textbf{Explanation faithfulness.}
TimeLitmus evaluates explanation faithfulness at three complementary levels.
Evidence validity checks whether cited variables satisfy the
domain-specific required, allowed, and forbidden evidence specification.
Linked faithfulness evaluates whether correct prediction and valid
evidence attribution occur jointly. Claim-conditioned faithfulness
uses controlled pairs to test whether a factor cited in the explanation
receives corresponding pair-level behavioral support. Citation Rate measures
whether the explanation mentions the manipulated factor, while
Claim-Supported Dependence Rate (CSDR) measures exact behavioral support among
cited factors. Additional metric decompositions are provided in Appendix~\ref{app:metric-definitions}.

\subsection{Domain Instantiations}
\label{sec:domain-instantiations}

\textbf{Finance.}
Finance uses four expected market-response categories:
\begin{equation}
\mathcal{S}_{F}
=
\{\mathrm{SU},\mathrm{WU},\mathrm{SD},\mathrm{WD}\},
\end{equation}
denoting strong upward, weak upward, strong downward, and weak downward
responses. Its rule system captures the interaction between event polarity
and the pre-event market state. Event semantics primarily determine response
direction, while temporal context modulates expected strength. For example, a
positive earnings surprise under a stable pre-event state supports a stronger
upward response than the same surprise following a substantial prior run-up,
where part of the information may already be reflected in price.

Finance controlled pairs instantiate this interaction from both directions.
Series-side interventions vary pre-event conditions such as trend,
volatility, or price state while preserving the financial event. Event-side
interventions vary mechanism-relevant event attributes while preserving the
same market history. Hard contrasts and explanation diagnostics extend the
same evidence specification to realistic, independently evaluated examples.

\textbf{Traffic.}
Traffic uses three ordered expected-response categories:
\begin{equation}
\mathcal{S}_{T}
=
\{s^{(T)}_1,s^{(T)}_2,s^{(T)}_3\},
\qquad
s^{(T)}_1 < s^{(T)}_2 < s^{(T)}_3.
\label{eq:traffic-response-space}
\end{equation}
Roadway incidents are aligned with nearby same-freeway, same-direction
sensors. Each instance combines an incident report with a pre-report traffic
history sampled before the report time; the report-time observation is
excluded from core prediction tasks. This construction preserves information
available before the reported incident response while avoiding direct access
to the target-period observation.

The Traffic rule system combines the pre-report temporal state with incident capacity-impact semantics.
Temporal factors include speed level, recent deceleration, existing congestion, and remaining deterioration capacity, while capacity-impact factors include shoulder or active-lane involvement, affected-lane count, and direct obstruction of the main traffic stream.
Series-side interventions vary the pre-report traffic state
while retaining the same incident description. Event-side interventions vary
incident severity or capacity impact while preserving the same traffic
history. This enables controlled evaluation of whether models respond to both
the incident and the temporal conditions under which it occurs.

Representative instances from both domains are provided in
Appendix~\ref{app:illustrative-instances}, including event text, serialized
time-series variables, response options, gold responses, and required
evidence.

\begin{figure}[t]
    \centering
    \includegraphics[
        width=0.94\linewidth,
        keepaspectratio
    ]{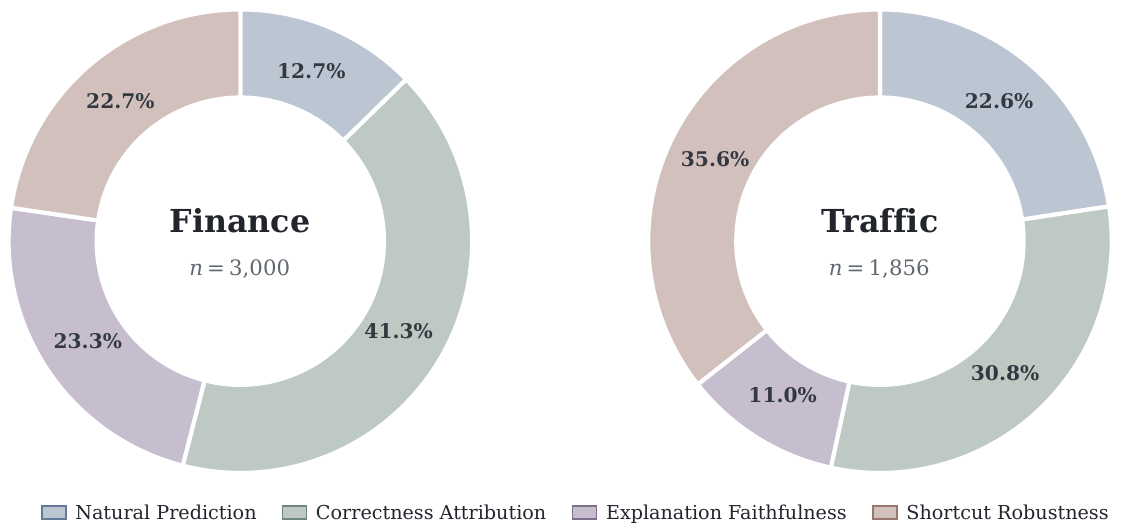}
    \caption{
    TimeLitmus composition across Finance and Traffic. The benchmark organizes
    4,856 evaluation records into Natural Prediction, Correctness Attribution,
    Explanation Faithfulness, and Shortcut Robustness.
    }
    \label{fig:3}
\end{figure}

\subsection{Construction and Quality Control}
\label{sec:rule-validation-quality}
\textbf{Rule-grounded controlled diagnostics.}
Controlled diagnostics require interventions with identifiable,
auditable, and reproducible expected effects. Following established
diagnostic evaluation, we use literature-grounded rules to isolate the
targeted event--series relationship and define expected changes or
invariances~\citep{gardner2020evaluating,ribeiro2020beyond,thrush2022winoground}.
Realized outcomes provide important empirical evidence, but also reflect
concurrent shocks and unobserved influences beyond the benchmark input;
we therefore use them as complementary external validation rather than
the sole basis of intervention-specific labels. This design combines
mechanism-level control with empirical grounding.

Rules are grounded in established finance and traffic-flow findings and
operationalized over frozen source-domain metadata
\citep{ball1968empirical,womack1996brokerage,
lighthill1955kinematic,knoop2009capacity}. Controlled instances undergo
deterministic reconstruction and intervention-isolation checks to verify
their labels, expected relations, and manipulated factors. This process
ensures that each retained pair changes the intended mechanism while
preserving the remaining model-visible context.

Two authors independently review constructed units using a predefined
quality-control checklist, achieving 98\% raw agreement. Disagreements are
adjudicated before benchmark freezing, and units that do not satisfy the
construction criteria are excluded. We additionally validate the evaluation
pipeline with oracle and adversarial mock outputs to detect unintended scorer
shortcuts and verify expected metric behavior.

Construction audits confirm that the benchmark rules and structured
diagnostics are reproducible from frozen source data. In Finance, an
observable rule pipeline recovers response direction with 99.79\% accuracy
and the complete four-way response with 85.35\% accuracy on covered
instances. In Traffic, every retained controlled relation passes deterministic
reconstruction and strict-transition checks, and all frozen Natural Prediction
and HPC labels are exactly reproducible from the post-report response
construction. Additional source-domain analyses are reported in
Appendix~\ref{app:finance-rule-validation} and
Appendix~\ref{app:traffic-rule-validation}.

\section{Experiments}
\label{sec:experiments}

The experiments are organized around three research questions:

\begin{itemize}[leftmargin=*,topsep=2pt,itemsep=1pt,parsep=0pt]
    \item \textbf{RQ1: Cross-modal integration.}
    Does natural-task accuracy reflect reliable integration of event semantics
    and pre-event time-series evidence?

    \item \textbf{RQ2: Explanation faithfulness.}
    Are the evidence factors cited in model explanations supported by the
    corresponding prediction behavior?

    \item \textbf{RQ3: Adaptation transfer.}
    Does natural-only adaptation transfer to controlled consistency,
    hard-pair behavior, and explanation-faithfulness diagnostics?
\end{itemize}

\subsection{Experimental Setup}
\label{sec:experimental-setup}

\textbf{Models and protocol.}
We evaluate ten LLMs: DeepSeek R1~\citep{guo2025deepseek}, DeepSeek v4 Flash~\citep{xu2026deepseek}, Gemini 3.5 Flash~\citep{gemini35flash}, MiniMax-M3~\citep{lai2026minimax}, Qwen-Plus~\citep{yang2025qwen3}, GPT-5.4~\citep{singh2025openai}, Claude Sonnet 4.6~\citep{claude_sonnet46}, GLM-5~\citep{zeng2026glm}, Qwen3.5-9B~\citep{qwen3.5}, and Qwen3.5-4B~\citep{qwen3.5}. 

We additionally train QLoRA~\citep{dettmers2023qlora} variants of Qwen3.5-4B and Qwen3.5-9B with multiple independent seeds. For each run, checkpoint selection uses held-out natural-prediction development performance.
All systems receive the same model-visible inputs: event text, serialized
time-series variables, response options, and output instructions. Controlled
endpoints and HPC arms are evaluated independently; pair identities, arm roles,
manipulated factors, expected transitions, private evidence annotations, and
answer keys remain hidden. Only the auxiliary relation probe presents both
scenarios jointly. We use deterministic decoding when available and score
invalid or unparsable outputs as incorrect. Manipulated-factor mappings and cited-factor annotations are manually
verified before claim-conditioned scoring.

\textbf{Reporting and uncertainty.}
Unless otherwise stated, results are percentages reported as
\textit{Finance / Traffic}. Confidence intervals are estimated by bootstrap
resampling over the native diagnostic unit: records for standalone tasks,
complete pairs for controlled and HPC diagnostics, and complete links for
linked diagnostics. QLoRA results are averaged across independently trained runs, with paired
Base--QLoRA effects computed on identical evaluation units. Full aggregation
and uncertainty details are provided in Appendix~\ref{app:cross-domain-aggregation} and Appendix~\ref{app:statistical-testing}.

\subsection{Metrics}
\label{sec:metrics}

Natural Accuracy measures full-input prediction accuracy. For each
expected-change pair, Counterfactual Pair Correctness (CF-PC) requires both
independently evaluated endpoints to match their benchmark-defined labels.
Because CF-PC is conjunctive, we additionally report
\begin{equation}
\Delta_{\mathrm{CF}}
=
\mathrm{CFPC}_{\mathrm{obs}}
-
\mathrm{CFPC}_{\mathrm{ind}},
\label{eq:delta-cf}
\end{equation}
where $\mathrm{CFPC}_{\mathrm{ind}}$ is the product of source- and target-arm
accuracies within the same domain and intervention side. Negative values
indicate lower pair consistency than expected from the endpoint accuracies
alone.

HPC Pair applies the both-endpoints-correct criterion to naturally occurring
hard contrasts. Evidence Recall measures coverage of required evidence
variables. Citation Rate measures how often a model cites the factor
manipulated by a controlled pair, while Claim-Supported Dependence Rate (CSDR)
measures how often such cited factors receive exact pair-level behavioral
support. The Auxiliary Relation probe evaluates explicit relation recognition
when both scenarios are presented jointly.

\subsection{Human Solvability}
\label{sec:human-solvability}

We conduct a matched human evaluation to assess whether the diagnostic
distinctions measured by TimeLitmus are recoverable from the provided
event--series inputs. Three independent annotators each complete 200
independently presented questions across Finance and Traffic, yielding 600
human judgments in total. The evaluation covers Natural Prediction,
series-side controlled interventions, and Hard Paired Contrasts under the same
model-visible input and independent-endpoint protocol used for LLM evaluation.
Each metric is computed separately for each annotator and then averaged across
annotators. Full protocol and per-annotator results are provided in
Appendix~\ref{app:human-solvability}.

\begin{table}[t]
\centering
\caption{
Human solvability on a matched 200-question evaluation per annotator.
Human results average three independent annotators, totaling 600 judgments.
The matched LLM envelope selects the strongest evaluated model separately for
each metric on the identical instances.
}
\label{tab:human-solvability}

\small
\setlength{\tabcolsep}{5.0pt}
\renewcommand{\arraystretch}{1.08}

\begin{tabular}{@{}llccc@{}}
\toprule
\textbf{Domain}
& \textbf{Evaluator}
& \textbf{Natural}
& \shortstack{\textbf{Series}\\\textbf{CF-PC}}
& \shortstack{\textbf{HPC}\\\textbf{Pair}} \\
\midrule

Finance
& Metric-wise Best LLM
& 55.0
& 20.0
& 25.0 \\

Finance
& \textbf{Human Avg.}
& \textbf{71.7}
& \textbf{53.3}
& \textbf{36.7} \\

\midrule

Traffic
& Metric-wise Best LLM
& 45.0
& 30.0
& 15.0 \\

Traffic
& \textbf{Human Avg.}
& \textbf{70.0}
& \textbf{46.7}
& \textbf{41.7} \\

\bottomrule
\end{tabular}
\end{table}

Human performance exceeds the matched LLM envelope across all evaluated
dimensions. The largest separation appears in Finance series-side CF-PC,
where humans achieve $53.3\%$, compared with $20.0\%$ for the strongest LLM
on the same instances. In Traffic, humans exceed the LLM envelope by
$25.0$ percentage points in Natural Accuracy and $26.7$ points in HPC Pair
Correctness. The consistent advantage across domains and diagnostic families
shows that TimeLitmus captures event--series distinctions that are accessible
to human reasoning but remain challenging for current LLMs.

\subsection{RQ1: Cross-Modal Integration}
\label{sec:cross-modal-results}
\textbf{Natural-task accuracy substantially overstates reliable cross-modal
integration.}
Table~\ref{tab:cross-modal-results} reveals a consistent separation between
natural prediction and controlled behavior. Natural Accuracy reaches 51.3\%
in Finance and 42.1\% in Traffic, but this performance does not translate into
reliable behavior under controlled changes.

All ten models exhibit negative Finance series-side
$\Delta_{\mathrm{CF}}$, ranging from $-16.4$ to $-6.3$ points, with all
corresponding 95\% bootstrap intervals below zero. Their pair correctness is
therefore lower than expected even after accounting for the difficulty of the
two endpoints. Event-side results are more heterogeneous, indicating that
sensitivity to event semantics and sensitivity to the pre-event temporal state
remain distinct capabilities.

\begin{table}[t]
\centering
\caption{
Cross-modal prediction and controlled consistency. Each cell reports
Finance / Traffic percentages. CF-PC requires both endpoints to be correct;
negative $\Delta_{\mathrm{CF}}$ indicates lower-than-expected pair consistency
under endpoint independence.
}
\label{tab:cross-modal-results}

\scriptsize
\setlength{\tabcolsep}{2.2pt}
\renewcommand{\arraystretch}{0.96}

\resizebox{\columnwidth}{!}{%
\begin{tabular}{@{}lcccccc@{}}
\toprule
& \textbf{Natural}
& \multicolumn{2}{c}{\textbf{Series-side}}
& \multicolumn{2}{c}{\textbf{Event-side}}
& \textbf{HPC} \\
\cmidrule(lr){3-4}
\cmidrule(lr){5-6}
\textbf{Model}
& \textbf{Acc.}
& \textbf{CF-PC}
& $\boldsymbol{\Delta_{\mathrm{CF}}}$
& \textbf{CF-PC}
& $\boldsymbol{\Delta_{\mathrm{CF}}}$
& \textbf{Pair} \\
\midrule

DeepSeek v4 Flash
& 48.7 / 36.9
& 8.0 / 18.4
& -8.7 / -0.9
& 17.5 / 33.3
& -3.3 / -7.1
& 16.7 / \textbf{11.7} \\

DeepSeek R1
& 44.5 / 41.0
& \textbf{13.7} / \textbf{21.3}
& \textbf{-6.3} / -3.0
& 18.8 / 26.7
& -5.5 / -8.0
& 16.1 / \textbf{11.7} \\

Gemini 3.5 Flash
& 42.1 / 35.2
& 7.0 / 8.8
& -15.9 / +1.0
& 37.5 / 6.7
& +0.8 / -12.0
& 9.4 / 0.0 \\

MiniMax-M3
& 47.1 / 37.1
& 11.0 / 11.8
& -7.2 / 0.0
& 25.0 / 56.7
& +1.8 / +0.7
& 13.9 / 6.7 \\

Qwen-Plus
& 41.1 / \textbf{42.1}
& 4.3 / 16.9
& -16.4 / -3.5
& \textbf{61.3} / 13.3
& -1.6 / -1.3
& 8.9 / 6.7 \\

GPT-5.4
& 44.7 / 34.5
& 3.0 / 8.8
& -10.8 / 0.0
& 5.0 / \textbf{63.3}
& -7.5 / \textbf{+4.2}
& 8.6 / 0.0 \\

Claude Sonnet 4.6
& 37.1 / 37.9
& 9.7 / 11.0
& -11.2 / +0.1
& 50.0 / 50.0
& \textbf{+2.7} / -1.0
& 15.0 / 1.7 \\

GLM-5
& \textbf{51.3} / 35.0
& 11.3 / 4.4
& -7.1 / +0.5
& 16.2 / 56.7
& -1.1 / +1.8
& \textbf{19.2} / 1.7 \\

Qwen3.5-9B
& 40.3 / 40.7
& 3.7 / 16.9
& -15.9 / -4.5
& 38.8 / 23.3
& -3.5 / -3.3
& 3.3 / \textbf{11.7} \\

Qwen3.5-4B
& 40.8 / 38.6
& 9.3 / 20.6
& -10.0 / \textbf{+2.0}
& 13.8 / 50.0
& -2.0 / -4.0
& 12.2 / 6.7 \\

\bottomrule
\end{tabular}%
}
\end{table}

HPC Pair Correctness is also consistently low, ranging from 3.3--19.2\% in
Finance and 0.0--11.7\% in Traffic. 
In contrast, models often perform
substantially better when the scenario relation is presented explicitly
in the Auxiliary Relation probe. 
This recognition--application separation
shows that identifying a contrast does not guarantee that the model will apply it consistently during
independent prediction.
On the matched Traffic controls, series-only input matches or exceeds full-input accuracy across models.
This reveals a strong shortcut pathway that
natural accuracy alone cannot distinguish from genuine event--series integration. 
Additional invariance and shortcut-control results are reported
in Appendix~\ref{app:shortcut-details}.

\subsection{RQ2: Explanation Faithfulness}
\label{sec:faithfulness-results}
\textbf{Evidence recognition and citation substantially exceed exact
behavioral support.}
Models can often identify task-relevant evidence, yet recognizing or citing
the correct factor does not ensure that predictions actually follow it.
Table~\ref{tab:claim-conditioned-faithfulness} directly evaluates this
claim--behavior alignment across Finance and Traffic.

A consistent gap emerges in both domains. Event-side factors are cited almost
universally, while Event CSDR varies substantially across models. The
separation is particularly pronounced for Traffic series-side interventions:
Citation exceeds 90\% for most models, whereas CSDR
remains below 22\%. Finance exhibits the same broader pattern, with behavioral
support remaining substantially lower than factor citation. Mentioning the
manipulated event or temporal factor therefore does not ensure that the
prediction actually depends on it.

\begin{table}[t]
\centering
\caption{
Claim-conditioned explanation faithfulness across Finance and Traffic.
Each cell reports Finance / Traffic percentages. Citation measures whether
the manipulated factor is cited, while CSDR measures exact pair-level
behavioral support among cited factors.
}
\label{tab:claim-conditioned-faithfulness}

\scriptsize
\setlength{\tabcolsep}{2.4pt}
\renewcommand{\arraystretch}{0.96}

\resizebox{\columnwidth}{!}{%
\begin{tabular}{@{}lcccc@{}}
\toprule
\textbf{Model}
& \shortstack{\textbf{Series}\\\textbf{Citation}}
& \shortstack{\textbf{Series}\\\textbf{CSDR}}
& \shortstack{\textbf{Event}\\\textbf{Citation}}
& \shortstack{\textbf{Event}\\\textbf{CSDR}} \\
\midrule

DeepSeek v4 Flash
& 22.3 / 94.1
& 16.4 / 18.8
& 100.0 / 100.0
& 17.5 / 33.3 \\

DeepSeek R1
& 11.0 / 91.9
& 30.3 / 21.6
& 97.5 / 100.0
& 19.2 / 26.7 \\

Gemini 3.5 Flash
& 17.7 / 99.3
& 17.0 / 8.9
& 100.0 / 100.0
& 37.5 / 6.7 \\

MiniMax-M3
& 54.7 / 99.3
& 12.8 / 11.1
& 100.0 / 100.0
& 25.0 / 56.7 \\

Qwen-Plus
& 22.7 / 98.5
& 8.8 / 17.2
& 100.0 / 100.0
& 61.3 / 13.3 \\

GPT-5.4
& 38.0 / 96.3
& 3.5 / 9.2
& 100.0 / 100.0
& 5.0 / 63.3 \\

Claude Sonnet 4.6
& 39.3 / 100.0
& 16.1 / 11.0
& 100.0 / 100.0
& 50.0 / 50.0 \\

GLM-5
& 53.3 / 100.0
& 16.2 / 4.4
& 100.0 / 100.0
& 16.2 / 56.7 \\

Qwen3.5-9B
& 13.3 / 58.8
& 5.0 / 15.0
& 98.8 / 66.7
& 39.2 / 30.0 \\

Qwen3.5-4B
& 16.0 / 27.2
& 14.6 / 13.5
& 96.2 / 36.7
& 14.3 / 63.6 \\

\bottomrule
\end{tabular}%
}
\end{table}

\begin{figure}[t]
    \centering
    \IfFileExists{4.pdf}{%
        \includegraphics[
            width=1\linewidth,
            height=0.35\textheight,
            keepaspectratio
        ]{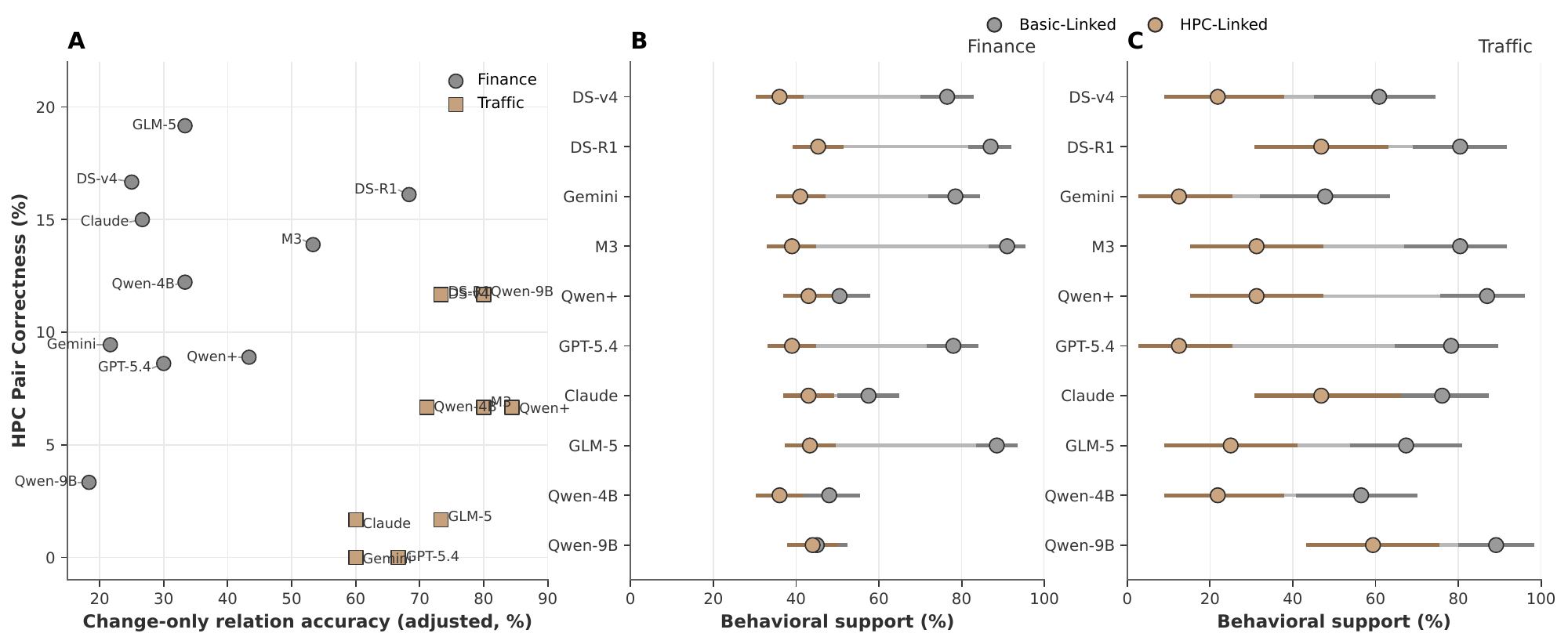}%
    }{%
        \fbox{\parbox[c][0.255\textheight][c]{0.96\textwidth}{%
        \centering
        \textbf{Diagnostic separations revealed by TimeLitmus}\\[4pt]
        Panel A: change-only Auxiliary Relation adjusted accuracy versus HPC Pair Correctness.\\
        Panels B--C: Basic-Linked versus HPC-Linked Behavioral Support with 95\% CIs.
        }}%
    }
    \caption{
    Diagnostic separations in TimeLitmus. Panel A contrasts relation recognition
    with HPC Pair Correctness; Panels B and C compare Basic- and HPC-Linked Behavioral Support across their
    respective linked sets, with 95\% bootstrap intervals.
    }
    \label{fig:recognition-faithfulness-gaps}
\end{figure}

The claim-conditioned results expose a gap between what explanations state
and what predictions behaviorally support.
Figure~\ref{fig:recognition-faithfulness-gaps} shows that this separation
persists under more challenging diagnostic conditions. Models recognize
explicit scenario relations more reliably than they apply them in independent
HPC prediction. Across their respective linked sets, Behavioral Support falls
from an average of 70.1\% / 72.4\% on Basic-linked records to
41.0\% / 30.9\% on HPC-linked records. Together, these results distinguish
evidence recognition, explanation generation, and behaviorally faithful
application as separate capabilities.

\begin{figure}[t]
    \centering
    \includegraphics[
        width=1\columnwidth,
        keepaspectratio
    ]{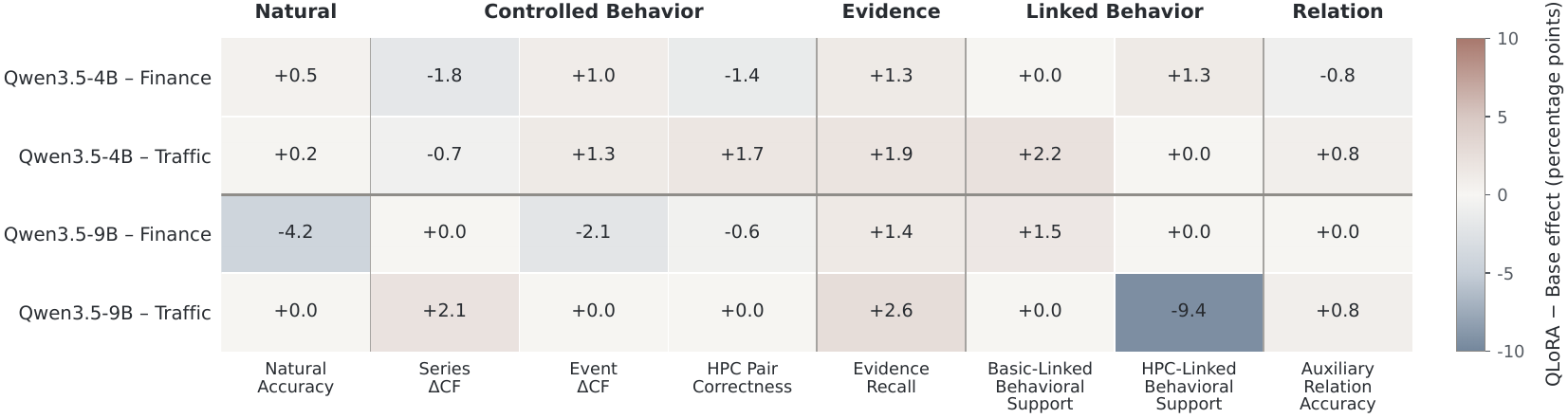}
    \caption{
    Paired QLoRA-minus-Base effects across scales and domains.
    Natural-only adaptation produces selective gains concentrated in evidence selection, distinct from transfer to controlled and hard-pair behavior.
    }
    \label{fig:paired-adaptation-effects}
\end{figure}

\textbf{Illustrative diagnostic case.}
Table~\ref{tab:finance-event-case} provides a concrete view of the diagnostic
separations identified in RQ1 and RQ2. The two endpoints use the same pre-event
series and differ only in the direction of the analyst recommendation.

\begin{table}[t]
\centering
\caption{
Representative Finance event-side pair for MiniMax-M3. TimeLitmus separates
endpoint correctness, controlled consistency, and claim-linked behavioral
support.
}
\label{tab:finance-event-case}

\scriptsize
\setlength{\tabcolsep}{3.5pt}
\renewcommand{\arraystretch}{1.12}

\begin{tabularx}{
\columnwidth
}{
@{}
>{\raggedright\arraybackslash}p{0.2\columnwidth}
>{\centering\arraybackslash}X
@{}
}
\toprule
\textbf{Component}
& \textbf{Endpoint contrast}
\\
\midrule

Shared series
& Same 30-day series; normalized price:
$100.0 \rightarrow 89.8$.
\\

Event
& Downgraded to Underperform
$\;\rightarrow\;$
Upgraded to 4.5-Stars
\\

Expected response
& Strong downward
$\;\rightarrow\;$
Strong upward
\\

Model prediction
& Strong downward
$\;\rightarrow\;$
Strong downward
\\

Explanation claim
& Negative downgrade
$\;\rightarrow\;$
Positive upgrade
\\

\midrule

Diagnostic profile
& One endpoint correct;\enspace
CF-PC $=0$;\enspace
Citation $=1$;\enspace
pair-level support $=0$.
\\

\bottomrule
\end{tabularx}
\end{table}

The model correctly distinguishes the negative downgrade from the positive
upgrade in its explanations, while producing the same response for both
endpoints. TimeLitmus resolves this output into a precise diagnostic profile:
the pair contains a correct endpoint and relevant factor citation, but not the
expected event-conditioned prediction transition. The example illustrates how
the benchmark separates correctness, controlled application, and
claim-supported behavior within a single evaluation unit.

\subsection{RQ3: Adaptation Transfer}
\label{sec:adaptation-results}
\textbf{Natural-only adaptation produces selective evidence-selection gains
but does not consistently transfer to controlled or hard-pair behavior.}
We train QLoRA variants of Qwen3.5-4B and Qwen3.5-9B on a disjoint
natural-prediction corpus constructed from the original Finance and Traffic
sources. Training instances contain model-visible natural-prediction inputs,
rule-grounded answers, evidence variables, and short rationales. Diagnostic
annotations and TimeLitmus evaluation answers are excluded from training and
checkpoint selection. The adaptation study uses
multiple independent seeds and development-selected checkpoints, while Base
and QLoRA outputs are aligned on identical diagnostic units for paired effect
estimation.

Figure~\ref{fig:paired-adaptation-effects} shows that natural-only adaptation
produces localized rather than broad diagnostic improvements.
The clearest positive effects concern evidence selection, whereas
series-side consistency, event-side consistency, HPC Pair Correctness,
and linked behavioral support show no consistent gains across model
scales and domains. Complete paired effects and confidence intervals
are reported in Appendix~\ref{app:paired-adaptation-results}.

\subsection{Probing Confidence-Level Input Sensitivity}

Behavioral metrics reveal whether adaptation improves
relation-consistent prediction, but they do not show whether it changes
which inputs support the model's confidence. We therefore conduct a
masking-based output-logit analysis on the open-weight Qwen3.5-4B and
Qwen3.5-9B models, providing an intermediate diagnostic layer between
evidence recognition and behavioral application.

For each model scale, we evaluate a shared set of 720 Finance and
Traffic instances spanning natural prediction, controlled diagnostics,
and evidence attribution. We mask the textual event, the historical
series, the recent temporal window, required-evidence inputs, and
forbidden cues, and measure the resulting change in the normalized
log-probability of the gold answer. Base and QLoRA models are evaluated
on identical instances, with QLoRA results averaged across three
independent seeds. Full definitions and stratified results are provided
in Appendix~\ref{app:logit-sensitivity}.

The probe reveals structured but selective shifts in input sensitivity.
For Qwen3.5-4B, adaptation increases confidence-level sensitivity to
both event and series inputs in Finance, whereas Traffic changes are
concentrated on the historical series and recent temporal window. The
9B model exhibits more localized effects. Importantly, these
confidence-level changes do not consistently translate into
improvements in CF-PC, HPC Pair Correctness, or linked behavioral
support. Together, the auxiliary relation and output-logit analyses
distinguish explicit relation recognition, confidence-level input
sensitivity, and relation-consistent behavioral application.

\subsection{Summary}
\label{sec:experiment-summary}

TimeLitmus reveals three capability separations that are obscured by
conventional evaluation. First, natural-task accuracy does not reliably
translate into controlled pair consistency: models can recognize explicit
scenario relations while applying them inconsistently during independent
prediction. Second, explanations frequently cite relevant event or temporal
factors without corresponding pair-level behavioral support, separating
factor mention from faithful application. Third, natural-only adaptation
produces selective improvements in evidence selection and, as revealed by
masking-based output-logit probing, can increase confidence-level sensitivity
to intended event, series, and evidence inputs. However, these shifts do not
consistently transfer to controlled consistency, hard-pair correctness, or
linked behavioral support. Together, these findings show that natural
prediction, confidence-level input sensitivity, controlled cross-modal
behavior, explanation faithfulness, and adaptation transfer constitute
distinct evaluation dimensions captured by TimeLitmus.

\section{Conclusion}
\label{sec:conclusion}

We introduced TimeLitmus, a multi-domain benchmark for evaluating
event-conditioned time-series prediction beyond surface accuracy. By combining
natural prediction with controlled series-side and event-side interventions,
hard paired contrasts, explanation-faithfulness diagnostics, and shortcut
controls, TimeLitmus tests whether model predictions are behaviorally
consistent with the joint use of textual events and historical time series.
Our experiments across Finance and Traffic show that natural accuracy can
coexist with weak pair-level consistency, relevant factors can be recognized
or cited without corresponding behavioral support, and natural-only adaptation
yields selective gains in evidence selection and confidence-level input
sensitivity without consistently closing these behavioral gaps. TimeLitmus
provides a unified diagnostic framework for evaluating event-conditioned
time-series models beyond surface correctness and plausible rationales.


\bibliographystyle{ACM-Reference-Format}
\bibliography{references}

\clearpage
\appendix
\setcounter{figure}{0}
\renewcommand{\thefigure}{A\arabic{figure}}

\section{Complete Definitions and Notation}
\label{app:notation}

\begin{table*}[t]
\centering
\caption{Notation used in the problem formulation.}
\label{tab:notation-summary}

\footnotesize
\setlength{\tabcolsep}{0pt}
\renewcommand{\arraystretch}{1.04}

\begin{tabular}{
@{}
>{\raggedright\arraybackslash}p{0.20\textwidth}
@{\hspace{6pt}}
>{\raggedright\arraybackslash}p{0.49\textwidth}
@{}
}
\toprule
\textbf{Symbol}
& \textbf{Definition}
\\
\midrule

$\mathcal{D}, d$
& Set of benchmark domains and a domain index.
\\

$L, p_d, \mathbf{x}_{\ell}^{(d)}, \mathbf{X}_d$
& History length, variable dimension, observation at time $\ell$, and
pre-event series.
\\

$t, \ell$
& Final pre-event index and a generic history index.
\\

$\mathcal{E}_d, E_d, \mathcal{Q}_d, Q_d$
& Event space, textual event, task-presentation space, and task presentation.
\\

$\mathcal{Y}_d, g_d, y_d$
& Response space, benchmark labeling function, and gold response.
\\

$M, \hat{y}_d, \mathcal{O}_d, O_d$
& Evaluated model, predicted response, explanation space, and generated
explanation.
\\

$\mathbb{I}[\cdot]$
& Indicator function.
\\

$\mathcal{N}_d, k, N_d^{\mathrm{Nat}}$
& Natural-prediction set, instance index, and number of natural instances.
\\

$m, i, b, N_{d,m}^{\mathrm{CF}}$
& Intervention side, controlled-pair index, endpoint index, and number of
controlled pairs.
\\

$Q_{d,m,i}, \tau_{d,m,i}$
& Shared task presentation and validated controlled transformation.
\\

$B_{d,m,i}, A_{d,m}^{(b)}$
& Both-endpoints-correct indicator and endpoint accuracy.
\\

$\mathrm{CF\text{-}PC}_{d,m},
\Delta_{\mathrm{CF},d,m}$
& Counterfactual Pair Correctness and residual pair consistency.
\\

$\mathcal{P}^{\mathrm{PI}}_d, j,
N_d^{\mathrm{PI}}, \mathrm{PI}_d$
& Label-preserving-pair set, index, count, and Prediction Invariance.
\\

$\mathcal{P}^{\mathrm{HPC}}_d, h,
N_d^{\mathrm{HPC}}, \mathrm{HPC\text{-}PC}_d$
& Hard-pair set, index, count, and HPC Pair Correctness.
\\

$\mathcal{F}_d, \phi_d$
& Canonical evidence-factor vocabulary and explanation-to-factor mapping.
\\

$\mathcal{G}_{d,m,i}, \mathcal{R}_{d,m,i}$
& Manipulated factor-family set and cited factor set.
\\

$C_{d,m,i}$
& Indicator that a manipulated factor is cited.
\\

$\mathrm{Citation}_{d,m},
\mathrm{CSDR}_{d,m}$
& Manipulated-factor citation rate and behaviorally supported dependence
rate among cited pairs.
\\

\bottomrule
\end{tabular}
\end{table*}

This appendix restores the full domain and intervention-side indices
suppressed in Section~\ref{sec:problem} and provides the standard metric
definitions omitted from the main text.

\subsection{Natural Prediction}
\label{app:natural-prediction}

Let $\mathcal{D}$ be the set of benchmark domains and
$d\in\mathcal{D}$ a domain index. For history length
$L\in\mathbb{N}_{+}$ and domain-specific variable dimension
$p_d\in\mathbb{N}_{+}$, the pre-event series is
\begin{equation}
\mathbf{X}_d
=
[\mathbf{x}_{t-L+1}^{(d)},\ldots,\mathbf{x}_{t}^{(d)}]
\in\mathbb{R}^{L\times p_d},
\label{eq:full-series}
\end{equation}
where $t$ is the final pre-event index and
$\mathbf{x}_{\ell}^{(d)}\in\mathbb{R}^{p_d}$ is the observation at time
$\ell$. Let $E_d\in\mathcal{E}_d$ be a textual event,
$Q_d\in\mathcal{Q}_d$ the task presentation, and
$\mathcal{Y}_d$ the semantic response space. The benchmark label and model
output follow Eq.~\eqref{eq:base-task}.

For domain $d$, the natural-prediction set is
\begin{equation}
\mathcal{N}_d
=
\left\{
(\mathbf{X}_{d,k},E_{d,k},Q_{d,k},y_{d,k})
\right\}_{k=1}^{N_d^{\mathrm{Nat}}},
\label{eq:natural-set}
\end{equation}
where $N_d^{\mathrm{Nat}}=|\mathcal{N}_d|$ and
$k\in\{1,\ldots,N_d^{\mathrm{Nat}}\}$ indexes an instance. Natural Accuracy
is
\begin{equation}
\mathrm{Acc}^{\mathrm{Nat}}_d
=
\frac{1}{N_d^{\mathrm{Nat}}}
\sum_{k=1}^{N_d^{\mathrm{Nat}}}
\mathbb{I}\!\left[\hat{y}_{d,k}=y_{d,k}\right].
\label{eq:natural-accuracy}
\end{equation}

\subsection{Controlled Expected-Change Pairs}
\label{app:controlled-pairs}

For domain $d$ and intervention side
$m\in\{\mathrm{series},\mathrm{event}\}$, let
$\mathcal{P}^{\mathrm{CF}}_{d,m}$ contain
$N_{d,m}^{\mathrm{CF}}=|\mathcal{P}^{\mathrm{CF}}_{d,m}|$ controlled
expected-change pairs. Pair
$i\in\{1,\ldots,N_{d,m}^{\mathrm{CF}}\}$ contains endpoints
$b\in\{0,1\}$ with a shared task presentation $Q_{d,m,i}$. Their labels are
\begin{equation}
y_{d,m,i}^{(b)}
=
g_d\!\left(
\mathbf{X}_{d,m,i}^{(b)},
E_{d,m,i}^{(b)}
\right).
\label{eq:full-pair-label}
\end{equation}
A validated transformation changes the designated modality while preserving
the other:
\begin{equation}
\left(
\mathbf{X}_{d,m,i}^{(1)},
E_{d,m,i}^{(1)}
\right)
=
\tau_{d,m,i}\!\left(
\mathbf{X}_{d,m,i}^{(0)},
E_{d,m,i}^{(0)}
\right),
\qquad
y_{d,m,i}^{(0)}\neq y_{d,m,i}^{(1)}.
\label{eq:full-pair-transformation}
\end{equation}
It changes $\mathbf{X}$ when $m=\mathrm{series}$ and $E$ when
$m=\mathrm{event}$. Each endpoint is evaluated independently:
\begin{equation}
(\hat{y}_{d,m,i}^{(b)},O_{d,m,i}^{(b)})
=
M\!\left(
\mathbf{X}_{d,m,i}^{(b)},
E_{d,m,i}^{(b)},
Q_{d,m,i}
\right).
\label{eq:full-pair-output}
\end{equation}

The pair-correctness indicator is
\begin{equation}
B_{d,m,i}
=
\prod_{b\in\{0,1\}}
\mathbb{I}\!\left[
\hat{y}_{d,m,i}^{(b)}
=
y_{d,m,i}^{(b)}
\right].
\label{eq:full-pair-indicator}
\end{equation}
Counterfactual Pair Correctness is
\begin{equation}
\mathrm{CF\text{-}PC}_{d,m}
=
\frac{1}{N_{d,m}^{\mathrm{CF}}}
\sum_{i=1}^{N_{d,m}^{\mathrm{CF}}}
B_{d,m,i}.
\label{eq:full-cf-pc}
\end{equation}
For endpoint $b$, the corresponding accuracy is
\begin{equation}
A_{d,m}^{(b)}
=
\frac{1}{N_{d,m}^{\mathrm{CF}}}
\sum_{i=1}^{N_{d,m}^{\mathrm{CF}}}
\mathbb{I}\!\left[
\hat{y}_{d,m,i}^{(b)}
=
y_{d,m,i}^{(b)}
\right].
\label{eq:full-endpoint-accuracy}
\end{equation}
Residual pair consistency is then
\begin{equation}
\Delta_{\mathrm{CF},d,m}
=
\mathrm{CF\text{-}PC}_{d,m}
-
A_{d,m}^{(0)}A_{d,m}^{(1)}.
\label{eq:full-delta-cf}
\end{equation}

In the main text, $N$, $Q_i$, $\tau_i$, $B_i$, $A^{(b)}$,
$\mathrm{CF\text{-}PC}$, and $\Delta_{\mathrm{CF}}$ denote the corresponding
quantities above within a fixed $(d,m)$ stratum.

\subsection{Label-Preserving Pairs}
\label{app:label-preserving}

Let $\mathcal{P}^{\mathrm{PI}}_d$ contain
$N_d^{\mathrm{PI}}=|\mathcal{P}^{\mathrm{PI}}_d|$ label-preserving pairs,
indexed by $j\in\{1,\ldots,N_d^{\mathrm{PI}}\}$. Their endpoints satisfy
$y_{d,j}^{(0)}=y_{d,j}^{(1)}$. Prediction Invariance is
\begin{equation}
\mathrm{PI}_d
=
\frac{1}{N_d^{\mathrm{PI}}}
\sum_{j=1}^{N_d^{\mathrm{PI}}}
\mathbb{I}\!\left[
\hat{y}_{d,j}^{(0)}
=
\hat{y}_{d,j}^{(1)}
\right].
\label{eq:prediction-invariance}
\end{equation}

\subsection{Hard Paired Contrasts}
\label{app:hard-pairs}

Let $\mathcal{P}^{\mathrm{HPC}}_d$ contain
$N_d^{\mathrm{HPC}}=|\mathcal{P}^{\mathrm{HPC}}_d|$ naturally occurring hard
pairs, indexed by $h\in\{1,\ldots,N_d^{\mathrm{HPC}}\}$. Each pair contains
similar observed endpoints satisfying
$y_{d,h}^{(0)}\neq y_{d,h}^{(1)}$. HPC Pair Correctness is
\begin{equation}
\mathrm{HPC\text{-}PC}_d
=
\frac{1}{N_d^{\mathrm{HPC}}}
\sum_{h=1}^{N_d^{\mathrm{HPC}}}
\prod_{b\in\{0,1\}}
\mathbb{I}\!\left[
\hat{y}_{d,h}^{(b)}
=
y_{d,h}^{(b)}
\right].
\label{eq:hpc-pair-correctness}
\end{equation}

\subsection{Claim-Conditioned Explanation Faithfulness}
\label{app:claim-conditioned}

Let $\mathcal{F}_d$ be the canonical evidence-factor vocabulary and
$\phi_d:\mathcal{O}_d\rightarrow2^{\mathcal{F}_d}$ the mapping from an
explanation to its cited factors. For controlled pair $i$ on side $m$, let
$\varnothing\neq\mathcal{G}_{d,m,i}\subseteq\mathcal{F}_d$ be the manipulated
factor-family set. The cited factors across the two explanations are
\begin{equation}
\mathcal{R}_{d,m,i}
=
\phi_d(O_{d,m,i}^{(0)})
\cup
\phi_d(O_{d,m,i}^{(1)}).
\label{eq:full-cited-set}
\end{equation}
The citation indicator is
\begin{equation}
C_{d,m,i}
=
\mathbb{I}\!\left[
\mathcal{R}_{d,m,i}
\cap
\mathcal{G}_{d,m,i}
\neq
\varnothing
\right].
\label{eq:full-citation-indicator}
\end{equation}
Citation Rate is
\begin{equation}
\mathrm{Citation}_{d,m}
=
\frac{1}{N_{d,m}^{\mathrm{CF}}}
\sum_{i=1}^{N_{d,m}^{\mathrm{CF}}}
C_{d,m,i}.
\label{eq:full-citation-rate}
\end{equation}
Claim-Supported Dependence Rate is
\begin{equation}
\mathrm{CSDR}_{d,m}
=
\frac{
\sum_{i=1}^{N_{d,m}^{\mathrm{CF}}}
C_{d,m,i}B_{d,m,i}
}{
\sum_{i=1}^{N_{d,m}^{\mathrm{CF}}}
C_{d,m,i}
},
\label{eq:full-csdr}
\end{equation}
provided that
$\sum_{i=1}^{N_{d,m}^{\mathrm{CF}}}C_{d,m,i}>0$.

In the main text, $\mathcal{G}_i$, $\mathcal{R}_i$, $C_i$,
$\mathrm{Citation}$, and $\mathrm{CSDR}$ denote the corresponding quantities
above within a fixed $(d,m)$ stratum.

\subsection{Shortcut Robustness}
\label{app:shortcut-robustness}

Partial-modality evaluations compare full-input prediction with event-only
and series-only prediction. Label-preserving controls test stability under
response-preserving transformations, whereas irrelevant-cue controls vary
presentation details or non-mechanistic metadata. These tests distinguish
joint event--series use from performance attainable through a single
modality or a superficial cue.

\subsection{Notation Summary}
\label{app:notation-summary}

Table~\ref{tab:notation-summary} summarizes the principal notation. Composite
subscripts combine the domain, intervention side, pair, endpoint, and
instance indices defined above.

\section{TimeLitmus Construction and Validation}
\label{app:timemub-construction}

This appendix documents the data sources, rule systems, construction pipeline,
quality-control procedures, and validation analyses used to build TimeLitmus.
The design follows a common principle across Finance and Traffic: domain
knowledge identifies task-relevant mechanisms, source data supports their
operationalization, and deterministic audits verify every released controlled
unit.

\subsection{Data Sources and Illustrative Instances}
\label{app:data-sources}

\paragraph{Finance.}
The Finance instantiation is based on FNSPID~\citep{dong2024fnspid}. We use
financial event texts and associated company or ticker metadata together with
historical equity variables, including normalized closing prices, daily
returns, and standardized trading volume. Rule-grounded expected responses are
constructed from event semantics and pre-event market states. Realized
post-event cumulative abnormal returns are used for aggregate rule-family
validation and auxiliary outcome analyses.

\paragraph{Traffic.}
The Traffic instantiation is based on TraffiDent~\citep{gou2026traffident}. We
use roadway incident descriptions containing freeway, direction, location,
postmile, and incident fields together with pre-report sensor histories of
speed, flow, and occupancy. Incidents are aligned with same-freeway,
same-direction sensors using the nearest sensor satisfying
$|\Delta\mathrm{PM}|\leq0.5$. Core prediction inputs contain only observations
available before the incident-report time.

\subsubsection{Illustrative Natural-Prediction Instances}
\label{app:illustrative-instances}

Figure~\ref{fig:timemub-examples} shows representative natural-prediction
instances from Finance and Traffic. The examples preserve the model-visible
structure of the released tasks: event text, serialized time-series variables,
response options, and a requested prediction with supporting evidence. Gold
labels and evidence annotations remain private during evaluation.

\begin{figure*}[t]
\small
\centering
\begin{minipage}[t]{0.48\textwidth}
\fbox{
\begin{minipage}{0.94\linewidth}
\textbf{Finance example.}

\textbf{Event:} Company A reported quarterly earnings above analyst
expectations.

\textbf{Pre-event time series:}

\texttt{normalized\_close}: [1.00, 1.02, 1.05, 1.08, 1.11, 1.13]

\texttt{return\_1d}: [0.01, 0.02, 0.03, 0.02, 0.02, 0.01]

\texttt{volume\_zscore}: [0.2, 0.5, 0.8, 1.1, 1.3, 1.0]

\textbf{Question:} What is the expected market reaction?

A. strong upward

B. weak upward

C. strong downward

D. weak downward

\textbf{Gold:} B. weak upward

\textbf{Required evidence:} positive event polarity; prior run-up or
pricing-in.
\end{minipage}
}
\end{minipage}
\hfill
\begin{minipage}[t]{0.48\textwidth}
\fbox{
\begin{minipage}{0.94\linewidth}
\textbf{Traffic example.}

\textbf{Event:} Incident on freeway F, northbound, near postmile P. A disabled
vehicle is blocking an active travel lane.

\textbf{Pre-report time series:}

\texttt{speed}: [64, 63, 62, 60, 57, 53]

\texttt{flow}: [1420, 1450, 1480, 1510, 1490, 1440]

\texttt{occupancy}: [0.11, 0.12, 0.13, 0.15, 0.18, 0.21]

\textbf{Question:} What is the expected traffic response?

A. minimal or no excess traffic disruption

B. moderate excess traffic disruption

C. severe excess traffic disruption

\textbf{Gold:} C. severe excess traffic disruption

\textbf{Required evidence:} recent speed decline; active-lane impact;
roadway-capacity reduction.
\end{minipage}
}
\end{minipage}
\caption{Illustrative TimeLitmus natural-prediction instances. Values are
shortened for readability. Released Finance prompts contain full serialized
pre-event financial series; released Traffic prompts contain 12 pre-report
time steps from $[-60,-5]$ minutes.}
\label{fig:timemub-examples}
\end{figure*}

\subsection{Benchmark Composition and Record Accounting}
\label{app:expanded-evaluation-record-accounting}

TimeLitmus reports benchmark scale using task-specific evaluation records.
Standalone tasks contribute one record, controlled diagnostics contribute
independently evaluated arms, linked diagnostics contribute one linked record,
and auxiliary probes contribute one relation-judgment record. Finance HPC
counts newly evaluated hard-target records because their source predictions
are already included in Natural Prediction.

Expanded-record accounting is used only for benchmark composition. All metrics
and confidence intervals are computed over their native statistical units,
including standalone instances, complete pairs, complete links, relation
probes, and matched-control units. Table~\ref{tab:expanded-evaluation-record-accounting}
reports the composition at the diagnostic-family level used in
Figure~\ref{fig:3}.

\begin{table}[t]
\caption{Family-level composition of TimeLitmus. Counts denote task-specific
diagnostic records.}
\label{tab:expanded-evaluation-record-accounting}
\centering
\apptablesingle
\setlength{\tabcolsep}{4pt}
\begin{tabular}{@{}lrrr@{}}
\toprule
\textbf{Diagnostic Family}
& \textbf{Finance}
& \textbf{Traffic}
& \textbf{Total} \\
\midrule
Natural Prediction
& 380 & 420 & 800 \\
Correctness Attribution
& 1,240 & 572 & 1,812 \\
Explanation Faithfulness
& 700 & 204 & 904 \\
Shortcut Robustness
& 680 & 660 & 1,340 \\
\midrule
\textbf{Total}
& \textbf{3,000}
& \textbf{1,856}
& \textbf{4,856} \\
\bottomrule
\end{tabular}
\end{table}

\subsection{Literature-Grounded Rule Development}
\label{app:literature-grounded-rules}

TimeLitmus follows a literature-first and data-validated rule development procedure. Candidate mechanisms are identified from established domain theory
and empirical findings, operationalized as deterministic predicates over event
attributes and pre-event time-series states, evaluated against source-domain
observations, and verified through construction audits and dual-review quality
control.

The literature supports the qualitative direction and mechanism structure of
the rules. In Finance, prior work establishes the information content of
earnings announcements, post-earnings price adjustment, analyst-recommendation
changes, and managerial guidance
\citep{ball1968empirical,bernard1989post,womack1996brokerage,
jegadeesh2004analyzing,anilowski2007does}. In Traffic, kinematic-wave and
cell-transmission models motivate state-dependent responses, while empirical
incident studies characterize capacity reductions associated with lane
blockage and driver behavior
\citep{lighthill1955kinematic,richards1956shock,daganzo1994cell,
knoop2008capacity,knoop2009capacity}. Benchmark-specific predicates,
thresholds, and response categories are determined through the source-data and
construction analyses below.

\subsection{Finance Rule System}
\label{app:finance-rules}

The Finance rule system captures the interaction between event semantics and
the pre-event market state. Event polarity determines the expected response
direction, while temporal context modulates expected strength. Favorable
earnings surprises, analyst upgrades, and upward guidance support upward
responses; unfavorable surprises, analyst downgrades, and downward guidance
support downward responses. Prior price movement, market anticipation, and
abnormal attention determine whether the expected response is strong or weak.

TimeLitmus uses four Finance response categories:
\[
\mathcal{S}_{F}
=
\{\mathrm{SU},\mathrm{WU},\mathrm{SD},\mathrm{WD}\},
\]
denoting strong upward, weak upward, strong downward, and weak downward
responses. The released rule inventory includes a core real-data layer and a
controlled diagnostic layer. The real-data layer uses earnings beats, analyst
upgrades, and earnings misses. The controlled layer additionally includes
analyst downgrades and guidance raises, enabling clean event-side and
series-side contrasts.

\paragraph{Series-side interventions.}
Finance series-side pairs hold the event fixed and replace the pre-event
sequence with another sequence satisfying a different market-state predicate.
The expected transition changes response strength while preserving event
polarity. This construction tests whether the same event is interpreted
differently under distinct temporal states.

\paragraph{Event-side interventions.}
Finance event-side pairs hold the normalized close, return, and volume history
fixed while changing a mechanism-relevant event attribute. The retained set
uses event changes that induce a strict expected-response transition under the
shared market history.

\paragraph{Invariance and shortcut controls.}
Response-preserving transformations keep the expected label unchanged.
Irrelevant-cue diagnostics manipulate option position, unsupported hints, or
non-mechanistic metadata. These controls test prediction stability and
forbidden-evidence avoidance.

\subsection{Traffic Rule System}
\label{app:traffic-rules}

The Traffic rule system is grounded in macroscopic traffic-flow theory and
empirical studies of incident-induced capacity reduction. Kinematic-wave and
cell-transmission models characterize traffic evolution as a function of the
pre-incident state, demand, density, and available roadway capacity
\citep{lighthill1955kinematic,richards1956shock,daganzo1994cell}. Empirical
incident studies further establish the effects of active-lane blockage, lane
loss, merging behavior, and driver distraction on effective roadway capacity
\citep{knoop2008capacity,knoop2009capacity}.

These mechanisms determine three ordered expected-response categories:
\begin{equation}
\begin{aligned}
A=s^{(T)}_1
&=\text{minimal or no excess traffic disruption},\\
B=s^{(T)}_2
&=\text{moderate excess traffic disruption},\\
C=s^{(T)}_3
&=\text{severe excess traffic disruption},
\end{aligned}
\end{equation}
with
\[
\operatorname{rank}(s^{(T)}_1)
<
\operatorname{rank}(s^{(T)}_2)
<
\operatorname{rank}(s^{(T)}_3).
\]

\paragraph{Event--sensor alignment.}
Each incident is matched to a same-freeway, same-direction sensor using the
nearest available sensor satisfying $|\Delta\mathrm{PM}|\leq0.5$. Core
prediction uses the pre-report window $[-60,-5]$ minutes sampled every five
minutes.

\paragraph{Series-side factors.}
Pre-report speed, flow, and occupancy operationalize the traffic state and its
remaining deterioration capacity. Relevant factors include baseline speed,
recent speed decline, persistent deceleration, existing congestion, flow and
occupancy patterns, and remaining room for further deterioration. The same
incident can therefore imply different responses under free-flowing,
deteriorating, or already congested conditions.

\paragraph{Event-side factors.}
Event-side rules operationalize incident-induced capacity loss through
active-lane involvement, affected-lane count, direct obstruction of the main
traffic stream, and roadway-capacity impact. Event-side pairs preserve the
historical series and modify only capacity-relevant incident attributes.

\paragraph{Strict pair construction.}
Only pairs with a unique strict expected transition are retained in the core
expected-change set. This yields an unambiguous pair-level gold relation and
prevents constant-label strategies from satisfying the diagnostic.

\paragraph{Invariance and shortcut controls.}
Traffic invariance pairs manipulate response-irrelevant information such as
option order, prompt metadata, or non-mechanistic context. They are used to
test both prediction stability and forbidden-cue avoidance.

\subsection{Shared Construction and Validation Framework}
\label{app:counterfactual-rule-validation}

Rule construction follows five stages:
\begin{enumerate}
    \item identify candidate mechanisms from domain theory and empirical
    literature;
    \item operationalize these mechanisms as deterministic predicates over
    event and pre-event time-series variables;
    \item evaluate the predicates against source-domain observations;
    \item conduct deterministic construction audits and dual-review quality
    control; and
    \item freeze only units satisfying the construction and review criteria.
\end{enumerate}

\paragraph{Deterministic reconstruction.}
For every controlled pair, the construction pipeline reconstructs the source
label, target label, intervention side, manipulated factor family, expected
relation, pair membership, and evidence target from private rule metadata. A
pair is retained only when the reconstructed label and relation match the
stored benchmark annotations.

\paragraph{Strict-transition validation.}
Expected-change pairs must exhibit the intended strict transition. Finance
pairs require a rule-consistent change in direction or strength. Traffic pairs
require a strict ordinal response transition. Label-preserving pairs are
verified against the corresponding invariance rule.

\paragraph{Intervention isolation.}
Each controlled unit is checked to ensure that the designated event or series
factor changes while the remaining model-visible task structure is preserved.
This audit connects the expected transition to a specific intervention family
and supports pair-level correctness attribution.

\paragraph{Manual-review protocol.}
Components containing at most 50 scientific units are reviewed in full.
Larger components use a stratified 10--15\% sample with at least 30 units per
component. Finance review is stratified by rule family and response label;
Traffic review is stratified by intervention side and response category.

Two authors independently review the same units using a predefined checklist
covering source lineage, input alignment, rule applicability, expected
response, intervention isolation, evidence annotations, and public--private
separation. Raw reviewer agreement is 98\%. Disagreements are adjudicated
before benchmark freezing, and units that do not satisfy the final criteria
are removed. Deterministic audits are applied to the complete controlled sets.

\paragraph{Oracle and scorer checks.}
We validate the evaluator using oracle and adversarial mock outputs spanning
controlled prediction, hard-pair, evidence, and linked-faithfulness metrics.
All observed outcomes match the predefined expected behavior.
Out-of-vocabulary evidence keys are rejected and receive no evidence credit.

\begin{table}[t]
\centering
\small
\caption{Evaluator sanity checks using oracle and adversarial mock outputs.}
\label{tab:scorer-unit-tests}

\setlength{\tabcolsep}{5pt}
\renewcommand{\arraystretch}{1.02}

\begin{tabular}{lcc}
\toprule
\textbf{Test case}
& \textbf{Expected}
& \textbf{Observed} \\
\midrule
Controlled-pair oracle
& 1.0 & 1.0 \\

Constant prediction on expected-change pairs
& 0.0 & 0.0 \\

Incorrect predictions on both endpoints
& 0.0 & 0.0 \\

Empty evidence set
& 0.0 & 0.0 \\

Out-of-vocabulary evidence key
& Rejected & Rejected \\

Identical relation for all HPC pairs
& 0.0 & 0.0 \\

Prediction--evidence oracle
& 1.0 & 1.0 \\

Prediction--evidence mismatch
& 0.0 & 0.0 \\
\bottomrule
\end{tabular}
\end{table}

Together, the literature grounding, deterministic reconstruction,
intervention-isolation checks, dual-review quality control, and scorer tests
establish a reproducible validation chain for the TimeLitmus diagnostics.

\subsection{Finance Rule Validation}
\label{app:finance-rule-validation}

Finance validation combines observable rule recoverability with aggregate
cumulative-abnormal-return contrasts.

\paragraph{Observable rule recoverability.}
A transparent pipeline using observable event attributes and pre-event regime
features recovers the stored rule-grounded labels on 1,444 covered instances.
Direction accuracy is 99.79\%, strength accuracy is 85.42\%, and complete
four-way accuracy is 85.35\%.

\begin{table}[t]
\centering
\caption{Finance observable rule recoverability. Pipeline coverage is measured
over the common 1,740-instance CAR-reconstructable subset.}
\label{tab:finance-rule-recoverability}
\apptablesingle
\small
\begin{tabular}{@{}lr@{}}
\toprule
\textbf{Audit stage or metric} & \textbf{Result} \\
\midrule
CAR-reconstructable instances & 1,740 \\
Observable-rule-covered instances & 1,444 \\
Observable pipeline coverage & 82.99\% \\
Direction accuracy & 99.79\% \\
Strength accuracy & 85.42\% \\
Four-way accuracy & 85.35\% \\
\bottomrule
\end{tabular}
\end{table}

\paragraph{Aggregate realized-response characterization.}
Candidate Finance rule families are evaluated using cumulative abnormal
returns over the primary $\mathrm{CAR}[0,5]$ window. For rule family $f$, the
direction-normalized contrast is
\begin{equation}
\Delta_{\mathrm{CAR},f}
=
d_f
\left(
\overline{\mathrm{CAR}}_{\mathrm{strong},f}
-
\overline{\mathrm{CAR}}_{\mathrm{weak},f}
\right),
\end{equation}
where $d_f=+1$ for upward event families and $d_f=-1$ for downward event
families. Positive values indicate that the strong-response group moves
further in the rule-expected direction.

\begin{table*}[t]
\caption{Direction-normalized realized-return contrasts for Finance rule
families under the primary $\mathrm{CAR}[0,5]$ window. Positive values
indicate that the strong-response group moves further in the rule-expected
direction.}
\label{tab:finance-rule-family-car-validation}
\centering
\apptablewide
\small
\begin{adjustbox}{max width=\textwidth,center}
\setlength{\tabcolsep}{4pt}
\begin{tabular}{@{}lcrrrrr@{}}
\toprule
\textbf{Rule Family}
& \textbf{Direction}
& \textbf{$n_{\mathrm{strong}}$}
& \textbf{$n_{\mathrm{weak}}$}
& \textbf{Mean}
& \textbf{Median}
& \textbf{95\% CI} \\
\midrule
Earnings beat
& Upward & 7,705 & 11,844
& $+0.935$ & $+0.416$
& $[+0.547,+1.319]$ \\
Analyst upgrade
& Upward & 10,606 & 30,356
& $+0.306$ & $+0.406$
& $[+0.143,+0.463]$ \\
Earnings miss
& Downward & 2,167 & 2,080
& $+0.403$ & $+0.004$
& $[+0.110,+1.119]$ \\
Analyst downgrade
& Downward & 225 & 284
& $+0.686$ & $+0.778$
& $[+0.102,+2.069]$ \\
Guidance raise
& Upward & 513 & 800
& $+1.009$ & $+0.586$
& $[+0.138,+2.274]$ \\
\bottomrule
\end{tabular}
\end{adjustbox}
\end{table*}

All retained rule families exhibit positive direction-normalized mean contrasts, with their bootstrap confidence intervals remaining above zero. These aggregate realized-return patterns provide empirical support for the rule-defined direction and strength distinctions used to construct both naturally occurring evaluation records and controlled diagnostic contrasts.

\subsection{Traffic Rule Validation}
\label{app:traffic-rule-validation}

Traffic validation combines complete controlled-relation reconstruction with
exact reproducibility of the frozen observed-response labels.

\paragraph{Formal reconstruction.}
For every retained controlled pair, the pipeline reconstructs the source
label, target label, intervention type, expected relation, manipulated factor
family, pair membership, and evidence target. The final controlled set
contains 166 strict relations, all of which satisfy the deterministic
reconstruction and transition checks.

\paragraph{Observed-response reproducibility.}
For the 540 Traffic Natural and Hard Paired Contrast records, the stored
response categories are fully reproducible from the frozen post-report
response-bucketing procedure. All 540 labels are recovered.

\subsection{Auxiliary Relation Probes}
\label{app:auxiliary-relation-probes}

The Auxiliary Relation probe evaluates whether a model can recognize the
expected relation between two scenarios when they are presented jointly. This
complements CF-PC, which evaluates the same relationship through independent
endpoint predictions.

\paragraph{Finance.}
Finance includes 120 relation-probe records derived from financial
event--series contrasts. Depending on the rule family, the model determines
whether the second scenario implies a stronger upward, weaker upward, stronger
downward, weaker downward, or approximately unchanged expected response.

\paragraph{Traffic.}
Traffic includes 120 relation-probe records derived from strict
counterfactual and invariance pairs: 60 series-side records, 30 event-side
records, and 30 invariance records. Each prompt asks whether the expected
traffic disruption in the second scenario should increase, decrease, or remain
approximately unchanged.

\subsection{Intervention-to-Evidence Mapping}
\label{app:intervention-evidence-mapping}

Claim-conditioned faithfulness uses a private mapping from each expected-change
pair to the factor family manipulated by that pair. The mapping is used to
score whether an explanation cites the manipulated factor and whether the
cited claim receives pair-level behavioral support.

Finance factor families include event polarity or direction, prior run-up,
prior decline, pricing-in or market anticipation, overbought or oversold
pre-event state, and abnormal volume or attention when applicable. Traffic
factor families include speed level, recent speed decline, persistent
deceleration, existing congestion, remaining deterioration capacity,
active-lane involvement, affected-lane count, direct obstruction, and
roadway-capacity impact.

A pair may map to an evidence family rather than a single lexical key. This
supports linguistically varied explanations while retaining deterministic
factor-level scoring.

\subsection{Prompt Integrity and Leakage Control}
\label{app:prompt-visibility}

All prompts expose only model-visible information: event text, serialized
numerical history, question, response options, and output instructions.
Private metadata, including rule identifiers, pair roles, manipulated factor
families, expected transitions, gold evidence, auxiliary relation labels, and
answer keys, are never shown to the evaluated model.

Counterfactual and invariance endpoints are presented independently. The model
does not observe the paired counterpart, pair identifier, arm role, expected
transition, or manipulated factor. This preserves the independent-prediction
setting required by correctness-attribution and claim-conditioned
faithfulness metrics.

\subsection{Controlled-Input Distribution Analysis}
\label{app:input-sanity-details}

We evaluate whether controlled construction preserves the observable
characteristics of natural event--series inputs beyond the intended
intervention. Finance compares 300 natural and 300 recombined prompts matched
by rule family, examining prediction distributions, invalid-output rates,
response balance, uncertainty language, and rationale citation rates.

Traffic tracks the provenance of natural, repaired, recombined, and
event-edited inputs and compares controlled and matched natural instances by
expected response, traffic regime, incident family, and capacity-impact
category. These analyses provide distribution-level checks that controlled
construction isolates the target mechanism without introducing conspicuous
marginal artifacts.

\subsection{Cross-Domain Aggregate Reporting}
\label{app:cross-domain-aggregation}

Cross-domain results use domain-level macro-averaging rather than pooling all
expanded records. This gives Finance and Traffic equal weight and prevents
expanded pair arms or linked records from being treated as independent
observations. Standalone tasks are resampled by record, pair-based tasks by
complete pair, linked-faithfulness tasks by complete link, and auxiliary
relation tasks by relation-probe record.

\subsection{Auxiliary Realized-Outcome Correspondence}
\label{app:traffic-realized-correspondence}

We additionally compare the pre-report Traffic rule taxonomy with independently
reconstructed post-report adjusted-response buckets. The analysis uses 240
unique native series-side endpoints, of which 232 have complete
matched-placebo baselines. Among these endpoints, 194 receive the same
three-class category under the pre-report rule system and the post-report
adjusted-response construction, corresponding to 83.62\% exact agreement.

\subsection{Statistical Testing and Confidence Intervals}
\label{app:statistical-testing}

Confidence intervals are estimated by bootstrap resampling over the native
diagnostic unit. Natural Prediction and standalone evidence metrics resample
individual records. CF-PC, HPC Pair, Citation, and CSDR resample complete
pairs. Linked-faithfulness metrics resample complete links. Matched shortcut
comparisons resample matched control units.

QLoRA-minus-Base effects are computed on aligned evaluation units and
bootstrapped as paired differences. Reported intervals are two-sided 95\%
bootstrap intervals. Model-level point estimates and diagnostic denominators
are computed from the same frozen evaluation outputs used in the main tables.

\section{Human Solvability Evaluation}
\label{app:human-solvability}

\paragraph{Standardized evaluation protocol.}
Three independent annotators who were not involved in benchmark construction
evaluated an identical matched subset from Finance and Traffic. 

Each annotator independently completed 200 questions, comprising
100 questions per domain and yielding 600 human judgments in total. The
evaluation covered Natural Prediction, series-side controlled interventions,
and Hard Paired Contrasts.

Annotators received the same information available to the evaluated models:
the event text, serialized time-series history, response options, and output
instructions. Paired endpoints were presented independently. Gold responses,
pair identities, endpoint roles, expected transitions, and private evidence
annotations were not included in the evaluation interface.

\paragraph{Scoring and agreement.}
Natural Accuracy is computed over individual instances. Series CF-PC and HPC
Pair Correctness require both independently evaluated endpoints of a pair to
be answered correctly. Each metric is first computed separately for each
annotator and then averaged across the three annotators. The annotators achieve
an overall raw agreement of $85.0\%$, indicating consistent application of the
shared evaluation criteria.

\paragraph{Comparison with LLM performance.}
Table~\ref{tab:human-solvability-full} compares average human performance with
the metric-wise best LLM results on both the identical human-evaluation subset
and the complete TimeLitmus benchmark. For each column, the best LLM score is
selected from all evaluated models for that metric.

\begin{table}[t]
\centering
\caption{
Human performance and metric-wise best LLM results on the matched subset and
the complete TimeLitmus benchmark. Human values are averaged across three
independent annotators, who achieve $85.0\%$ overall raw agreement.
All values are percentages.
}
\label{tab:human-solvability-full}

\small
\setlength{\tabcolsep}{4.2pt}
\renewcommand{\arraystretch}{1.08}

\begin{tabular}{@{}llccc@{}}
\toprule
\textbf{Domain}
& \textbf{Evaluator}
& \textbf{Natural}
& \shortstack{\textbf{Series}\\\textbf{CF-PC}}
& \shortstack{\textbf{HPC}\\\textbf{Pair}} \\
\midrule

\multirow{3}{*}{Finance}
& Best LLM (full)
& 51.3
& 13.7
& 19.2 \\

& Best LLM (matched)
& 55.0
& 20.0
& 25.0 \\

& \textbf{Human Avg.}
& \textbf{71.7}
& \textbf{53.3}
& \textbf{36.7} \\

\midrule

\multirow{3}{*}{Traffic}
& Best LLM (full)
& 42.1
& 21.3
& 11.7 \\

& Best LLM (matched)
& 45.0
& 30.0
& 15.0 \\

& \textbf{Human Avg.}
& \textbf{70.0}
& \textbf{46.7}
& \textbf{41.7} \\

\bottomrule
\end{tabular}
\end{table}

Human performance is higher than the metric-wise best LLM results across both
domains and all evaluated diagnostic dimensions. On the identical matched
subset, the largest Finance separation appears in series-side CF-PC, where
humans achieve $53.3\%$, compared with $20.0\%$ for the best LLM. In Traffic,
humans achieve $70.0\%$ Natural Accuracy and $41.7\%$ HPC Pair Correctness,
exceeding the matched best-LLM results by $25.0$ and $26.7$ percentage points,
respectively.

The complete-benchmark results exhibit the same pattern. Human performance
exceeds the full-benchmark best LLM by $39.6$ percentage points on Finance
series-side CF-PC and by $30.0$ points on Traffic HPC Pair Correctness.
Together with the high agreement across independent annotators, these results
show that TimeLitmus captures coherent and human-accessible event--series
distinctions while exposing substantial limitations in current LLMs'
cross-modal prediction and paired consistency.

\section{Experimental Details and Complete Results}
\label{app:experiment-details}

This appendix provides the complete metric definitions, model configurations,
evaluation protocol, and diagnostic results supporting the main analysis.
TimeLitmus reports each capability separately rather than collapsing natural
prediction, controlled behavior, evidence use, and relation recognition into a
single score.

\subsection{Metrics and Statistical Units}
\label{app:metric-definitions}

\paragraph{Natural Prediction.}
For a set of $N$ natural-prediction records, Natural Accuracy is
\begin{equation}
\mathrm{Acc}_{\mathrm{nat}}
=
\frac{1}{N}
\sum_{i=1}^{N}
\mathbb{I}[\hat{y}_i=y_i],
\end{equation}
where $\mathbb{I}[\cdot]$ is the indicator function.

\paragraph{Controlled endpoints and CF-PC.}
For a fixed domain and intervention side, let $G$ denote the set of
expected-change pairs. For pair $i\in G$, define endpoint-correctness indicators
\begin{equation}
a_i=\mathbb{I}[\hat{y}_{i}^{(0)}=y_{i}^{(0)}],
\qquad
b_i=\mathbb{I}[\hat{y}_{i}^{(1)}=y_{i}^{(1)}].
\end{equation}
Observed Counterfactual Pair Correctness is
\begin{equation}
\mathrm{CF\text{-}PC}_{\mathrm{obs}}
=
\frac{1}{|G|}
\sum_{i\in G} a_i b_i.
\end{equation}
Let $\hat{p}_0$ and $\hat{p}_1$ be the two endpoint accuracies in the same
domain and intervention-side stratum. The endpoint-independence baseline and
residual are
\begin{equation}
\mathrm{CF\text{-}PC}_{\mathrm{ind}}
=
\hat{p}_0\hat{p}_1,
\qquad
\Delta_{\mathrm{CF}}
=
\mathrm{CF\text{-}PC}_{\mathrm{obs}}
-
\mathrm{CF\text{-}PC}_{\mathrm{ind}}.
\end{equation}
Negative $\Delta_{\mathrm{CF}}$ indicates that exact pair consistency is lower
than expected from the two endpoint accuracies alone.

\paragraph{Hard Paired Contrasts.}
HPC Pair applies the both-endpoints-correct criterion to naturally occurring
hard contrasts. Finance HPC Target Accuracy measures accuracy over the 360
newly evaluated hard-target records whose source predictions are already
included in Natural Prediction. Traffic HPC Arm Accuracy measures accuracy
over 120 independently evaluated endpoints from 60 hard pairs. HPC Pair is the
common pair-level metric across domains.

\paragraph{Evidence validity.}
For record $i$, let $R_i$, $A_i$, $F_i$, and $P_i$ denote the required,
allowed, forbidden, and model-cited evidence-factor sets. Evidence precision
and required-evidence recall are
\begin{equation}
\mathrm{Prec}^{E}_i
=
\frac{|P_i\cap(R_i\cup A_i)|}{\max(1,|P_i|)},
\qquad
\mathrm{Rec}^{E}_i
=
\frac{|P_i\cap R_i|}{|R_i|}.
\end{equation}
Evidence F1 is the harmonic mean of precision and recall, with value zero when
both are zero. The forbidden-hit indicator is
\begin{equation}
\mathrm{FH}_i
=
\mathbb{I}[P_i\cap F_i\neq\varnothing].
\end{equation}
Aggregate evidence metrics average the corresponding record-level quantities.

\paragraph{Claim-conditioned faithfulness.}
For controlled pair $i$, let $C_i=1$ indicate that the explanations cite the
manipulated factor family and let $B_i=1$ indicate exact CF-PC success.
Citation Rate and Claim-Supported Dependence Rate are
\begin{equation}
\mathrm{Citation}
=
\frac{1}{N}
\sum_{i=1}^{N}C_i,
\qquad
\mathrm{CSDR}
=
\frac{\sum_{i=1}^{N}C_iB_i}
{\sum_{i=1}^{N}C_i}.
\end{equation}
CSDR is reported when at least one pair cites the manipulated factor. Citation
measures factor mention, while CSDR measures exact pair-level behavioral
support among cited pairs.

\paragraph{Linked Behavioral Support.}
Each linked diagnostic associates a prediction record with a corresponding
evidence or contrastive explanation record. Behavioral Support is the
prediction-correct indicator on the linked set. Basic-linked and HPC-linked
records are reported separately to measure support under standard and
hard-contrast conditions.

\paragraph{Auxiliary Relation probe.}
The relation probe presents two scenarios jointly and asks whether the expected
response should increase, decrease, or remain unchanged. We report raw relation
accuracy and majority-adjusted accuracy:
\begin{equation}
\mathrm{RelAcc}_{\mathrm{adj}}
=
\frac{
\mathrm{RelAcc}
-
\mathrm{RelAcc}_{\mathrm{maj}}
}{
1-\mathrm{RelAcc}_{\mathrm{maj}}
}.
\end{equation}
The primary recognition--application comparison uses change-only records,
which directly test directional transition recognition. Invariance records are
retained as a complementary consistency check.

All confidence intervals are estimated over each metric's native statistical
unit, following Appendix~\ref{app:statistical-testing}.

\subsection{Models and Evaluation Protocol}
\label{app:model-prompting-details}

All systems receive the same task prompts and output instructions.
Deterministic decoding is used when supported. Open-weight Qwen models are
evaluated from frozen local copies of the corresponding repositories, and
their adapted variants receive the same evaluation prompts as the base
models. QLoRA results are averaged across three independently trained runs,
with the best development checkpoint selected within each run.

Each model is evaluated on 3,000 Finance diagnostic records and 1,856 Traffic
diagnostic records under the benchmark accounting described in
Appendix~\ref{app:expanded-evaluation-record-accounting}. Controlled endpoints
and HPC arms are presented independently. Pair identities, arm roles,
manipulated factors, expected transitions, private evidence annotations, and
answer keys remain hidden from the evaluated models.

\begin{table}[t]
\caption{Models, exact identifiers, and evaluation forms used in the
experiments.}
\label{tab:model-identifiers}
\centering
\apptablesingle
\small

\begin{adjustbox}{max width=\columnwidth,center}
\setlength{\tabcolsep}{6pt}
\begin{tabular}{@{}llc@{}}
\toprule
\textbf{Model}
& \textbf{Exact identifier}
& \textbf{Evaluation form} \\
\midrule

DeepSeek v4 Flash
& \texttt{deepseek-v4-flash}
& API \\

DeepSeek R1
& \texttt{deepseek-r1-250528}
& API \\

Gemini 3.5 Flash
& \texttt{gemini-3.5-flash}
& API \\

MiniMax-M3
& \texttt{MiniMax-M3}
& API \\

Qwen-Plus
& \texttt{qwen-plus}
& API \\

GPT-5.4
& \texttt{gpt-5.4}
& API \\

Claude Sonnet 4.6
& \texttt{claude-sonnet-4-6}
& API \\

GLM-5
& \texttt{glm-5}
& API \\

Qwen3.5-4B
& \texttt{Qwen/Qwen3.5-4B}
& Local \\

Qwen3.5-9B
& \texttt{Qwen/Qwen3.5-9B}
& Local \\

Qwen3.5-4B-QLoRA
& \texttt{Qwen/Qwen3.5-4B} + QLoRA
& Three-run QLoRA mean \\

Qwen3.5-9B-QLoRA
& \texttt{Qwen/Qwen3.5-9B} + QLoRA
& Three-run QLoRA mean \\

\bottomrule
\end{tabular}
\end{adjustbox}
\end{table}

\subsection{Complete Cross-Modal Prediction Results}
\label{app:behavioral-details}

Table~\ref{tab:app-behavioral} reports natural accuracy, endpoint accuracy,
pair correctness, endpoint-independence baselines, residual pair consistency,
and HPC performance. The complete profile shows that natural-task performance
and controlled consistency capture distinct aspects of model behavior.

\begin{table*}[t]
\centering
\caption{Complete behavioral results. Cells report Finance / Traffic
percentages. S and E denote series-side and event-side interventions. Adapted
Qwen variants are reported separately from the unadapted cross-model
comparison.}
\label{tab:app-behavioral}
\scriptsize
\setlength{\tabcolsep}{2.2pt}
\resizebox{\textwidth}{!}{%
\begin{tabular}{lccccccccccc}
\toprule
\textbf{Model}
& \textbf{Natural}
& \textbf{S Arm}
& \textbf{S CF-PC}
& \textbf{S Ind.}
& \textbf{S $\Delta$}
& \textbf{E Arm}
& \textbf{E CF-PC}
& \textbf{E Ind.}
& \textbf{E $\Delta$}
& \textbf{HPC Arm}
& \textbf{HPC Pair} \\
\midrule
DeepSeek v4 Flash & 48.7 / 36.9 & 41.0 / 46.3 & 8.0 / 18.4 & 16.7 / 19.3 & -8.7 / -0.9 & 45.6 / 66.7 & 17.5 / 33.3 & 20.8 / 40.4 & -3.3 / -7.1 & 42.2 / 40.8 & 16.7 / 11.7 \\
DeepSeek R1 & 44.5 / 41.0 & 44.7 / 49.6 & 13.7 / 21.3 & 19.9 / 24.3 & -6.3 / -3.0 & 49.4 / 63.3 & 18.8 / 26.7 & 24.3 / 34.7 & -5.5 / -8.0 & 47.8 / 37.5 & 16.1 / 11.7 \\
Gemini 3.5 Flash & 42.1 / 35.2 & 47.8 / 43.0 & 7.0 / 8.8 & 22.9 / 7.8 & -15.9 / +1.0 & 60.6 / 51.7 & 37.5 / 6.7 & 36.7 / 18.7 & +0.8 / -12.0 & 45.3 / 34.2 & 9.4 / 0.0 \\
MiniMax-M3 & 47.1 / 37.1 & 42.7 / 44.5 & 11.0 / 11.8 & 18.2 / 11.8 & -7.2 / -0.0 & 48.1 / 76.7 & 25.0 / 56.7 & 23.2 / 56.0 & +1.8 / +0.7 & 44.7 / 37.5 & 13.9 / 6.7 \\
Qwen-Plus & 41.1 / 42.1 & 45.5 / 45.2 & 4.3 / 16.9 & 20.7 / 20.4 & -16.4 / -3.5 & 79.4 / 46.7 & 61.3 / 13.3 & 62.8 / 14.7 & -1.6 / -1.3 & 46.1 / 38.3 & 8.9 / 6.7 \\
GPT-5.4 & 44.7 / 34.5 & 37.2 / 43.4 & 3.0 / 8.8 & 13.8 / 8.8 & -10.8 / +0.0 & 35.6 / 78.3 & 5.0 / 63.3 & 12.5 / 59.1 & -7.5 / +4.2 & 44.2 / 32.5 & 8.6 / 0.0 \\
Claude Sonnet 4.6 & 37.1 / 37.9 & 46.0 / 43.0 & 9.7 / 11.0 & 20.9 / 10.9 & -11.2 / +0.1 & 68.8 / 73.3 & 50.0 / 50.0 & 47.3 / 51.0 & +2.7 / -1.0 & 41.1 / 34.2 & 15.0 / 1.7 \\
GLM-5 & 51.3 / 35.0 & 43.0 / 40.8 & 11.3 / 4.4 & 18.5 / 3.9 & -7.1 / +0.5 & 41.9 / 75.0 & 16.2 / 56.7 & 17.3 / 54.9 & -1.1 / +1.8 & 48.6 / 34.2 & 19.2 / 1.7 \\
Qwen3.5-4B & 40.8 / 38.6 & 44.0 / 46.7 & 9.3 / 20.6 & 19.4 / 18.6 & -10.0 / +2.0 & 40.0 / 75.0 & 13.8 / 50.0 & 15.8 / 54.0 & -2.0 / -4.0 & 41.9 / 37.5 & 12.2 / 6.7 \\
Qwen3.5-4B-QLoRA & 41.3 / 38.8 & 43.0 / 47.1 & 6.7 / 19.9 & 18.4 / 18.5 & -11.8 / +1.4 & 38.8 / 76.7 & 13.8 / 53.3 & 14.8 / 56.0 & -1.0 / -2.7 & 41.0 / 40.0 & 10.8 / 8.3 \\
Qwen3.5-9B & 40.3 / 40.7 & 44.3 / 48.5 & 3.7 / 16.9 & 19.6 / 21.4 & -15.9 / -4.5 & 65.0 / 56.7 & 38.8 / 23.3 & 42.2 / 26.7 & -3.5 / -3.3 & 42.9 / 45.0 & 3.3 / 11.7 \\
Qwen3.5-9B-QLoRA & 36.1 / 40.7 & 44.0 / 47.1 & 3.3 / 17.6 & 19.3 / 20.0 & -16.0 / -2.3 & 65.6 / 56.7 & 37.5 / 23.3 & 43.1 / 26.7 & -5.6 / -3.3 & 41.0 / 42.5 & 2.8 / 11.7 \\
\bottomrule
\end{tabular}}
\end{table*}

Table~\ref{tab:app-delta-ci} reports pair-bootstrap confidence intervals for
$\Delta_{\mathrm{CF}}$. All unadapted Finance series-side intervals remain
below zero, showing that the pair-consistency gap persists after accounting
for endpoint difficulty.

\begin{table*}[t]
\centering
\caption{$\Delta_{\mathrm{CF}}$ in percentage points with 95\%
pair-bootstrap confidence intervals.}
\label{tab:app-delta-ci}

\scriptsize
\setlength{\tabcolsep}{3.2pt}
\renewcommand{\arraystretch}{1.0}

\begin{adjustbox}{width=0.8\textwidth,center}
\begin{tabular}{@{}lcccc@{}}
\toprule
\textbf{Model}
& \textbf{Finance Series $\Delta$}
& \textbf{Traffic Series $\Delta$}
& \textbf{Finance Event $\Delta$}
& \textbf{Traffic Event $\Delta$} \\
\midrule
DeepSeek v4 Flash & -8.7 [-11.1, -6.2] & -0.9 [-4.8, +3.0] & -3.3 [-8.5, +2.5] & -7.1 [-12.7, -1.8] \\
DeepSeek R1 & -6.3 [-9.0, -3.6] & -3.0 [-7.1, +1.1] & -5.5 [-10.6, -0.3] & -8.0 [-14.8, -1.9] \\
Gemini 3.5 Flash & -15.9 [-18.0, -13.5] & +1.0 [-0.9, +2.9] & +0.8 [-4.4, +6.2] & -12.0 [-19.9, -2.9] \\
MiniMax-M3 & -7.2 [-9.8, -4.6] & -0.0 [-2.9, +2.6] & +1.8 [-3.7, +7.0] & +0.7 [-3.8, +5.6] \\
Qwen-Plus & -16.4 [-18.3, -14.2] & -3.5 [-7.5, +0.5] & -1.6 [-4.6, +1.5] & -1.3 [-8.7, +4.7] \\
GPT-5.4 & -10.8 [-12.9, -8.6] & +0.0 [-2.6, +2.1] & -7.5 [-11.5, -2.9] & +4.2 [+0.0, +10.0] \\
Claude Sonnet 4.6 & -11.2 [-13.6, -8.7] & +0.1 [-2.8, +2.8] & +2.7 [-2.0, +7.5] & -1.0 [-5.8, +4.2] \\
GLM-5 & -7.1 [-9.7, -4.6] & +0.5 [-1.1, +1.7] & -1.1 [-6.3, +4.1] & +1.8 [-4.0, +8.3] \\
Qwen3.5-9B & -15.9 [-18.1, -13.8] & -4.5 [-8.2, -0.5] & -3.5 [-8.0, +1.5] & -3.3 [-10.3, +3.9] \\
Qwen3.5-4B & -10.0 [-12.5, -7.4] & +2.0 [-1.4, +5.4] & -2.0 [-7.1, +3.3] & -4.0 [-8.3, +0.0] \\
\bottomrule
\end{tabular}
\end{adjustbox}
\end{table*}

\subsection{Relation Recognition and Independent Application}
\label{app:relation-probe-details}

The Auxiliary Relation probe evaluates explicit recognition of the
benchmark-defined relation when two scenarios are shown jointly. Change-only
scores isolate directional transition recognition and are compared with
independent HPC prediction in the main analysis.

\begin{table*}[t]
\centering
\caption{Auxiliary relation-probe results, reported as Finance / Traffic
percentages. Change-only majority baselines are 33.3\% in Finance and 50.0\%
in Traffic.}
\label{tab:app-relation}

\scriptsize
\setlength{\tabcolsep}{3pt}

\resizebox{0.94\textwidth}{!}{%
\begin{tabular}{@{}lccccccc@{}}
\toprule
\textbf{Model}
& \textbf{All Raw}
& \textbf{All Adj.}
& \textbf{Change Raw}
& \textbf{Change Adj.}
& \textbf{Event}
& \textbf{Series}
& \textbf{Invariance} \\
\midrule
DeepSeek v4 Flash & 62.5 / 90.0 & 50.0 / 84.0 & 50.0 / 86.7 & 25.0 / 73.3 & 66.7 / 93.3 & 41.7 / 83.3 & 100.0 / 100.0 \\
DeepSeek R1 & 84.2 / 90.0 & 78.9 / 84.0 & 78.9 / 86.7 & 68.3 / 73.3 & 70.0 / 100.0 & 83.3 / 80.0 & 100.0 / 100.0 \\
Gemini 3.5 Flash & 60.8 / 85.0 & 47.8 / 76.0 & 47.8 / 80.0 & 21.7 / 60.0 & 83.3 / 100.0 & 30.0 / 70.0 & 100.0 / 100.0 \\
MiniMax-M3 & 76.7 / 92.5 & 68.9 / 88.0 & 68.9 / 90.0 & 53.3 / 80.0 & 90.0 / 100.0 & 58.3 / 85.0 & 100.0 / 100.0 \\
Qwen-Plus & 71.7 / 94.2 & 62.2 / 90.7 & 62.2 / 92.2 & 43.3 / 84.4 & 90.0 / 100.0 & 48.3 / 88.3 & 100.0 / 100.0 \\
GPT-5.4 & 65.0 / 87.5 & 53.3 / 80.0 & 53.3 / 83.3 & 30.0 / 66.7 & 90.0 / 100.0 & 35.0 / 75.0 & 100.0 / 100.0 \\
Claude Sonnet 4.6 & 63.3 / 85.0 & 51.1 / 76.0 & 51.1 / 80.0 & 26.7 / 60.0 & 90.0 / 100.0 & 31.7 / 70.0 & 100.0 / 100.0 \\
GLM-5 & 66.7 / 90.0 & 55.6 / 84.0 & 55.6 / 86.7 & 33.3 / 73.3 & 73.3 / 100.0 & 46.7 / 80.0 & 100.0 / 100.0 \\
Qwen3.5-4B & 66.7 / 89.2 & 55.6 / 82.7 & 55.6 / 85.6 & 33.3 / 71.1 & 80.0 / 96.7 & 43.3 / 80.0 & 100.0 / 100.0 \\
Qwen3.5-4B-QLoRA & 65.8 / 90.0 & 54.4 / 84.0 & 54.4 / 86.7 & 31.7 / 73.3 & 80.0 / 96.7 & 41.7 / 81.7 & 100.0 / 100.0 \\
Qwen3.5-9B & 59.2 / 92.5 & 45.6 / 88.0 & 45.6 / 90.0 & 18.3 / 80.0 & 60.0 / 100.0 & 38.3 / 85.0 & 100.0 / 100.0 \\
Qwen3.5-9B-QLoRA & 59.2 / 93.3 & 45.6 / 89.3 & 45.6 / 91.1 & 18.3 / 82.2 & 56.7 / 100.0 & 40.0 / 86.7 & 100.0 / 100.0 \\
\bottomrule
\end{tabular}%
}
\end{table*}

\begin{figure*}[t]
\centering
\IfFileExists{A2.pdf}{%
    \includegraphics[width=0.98\textwidth]{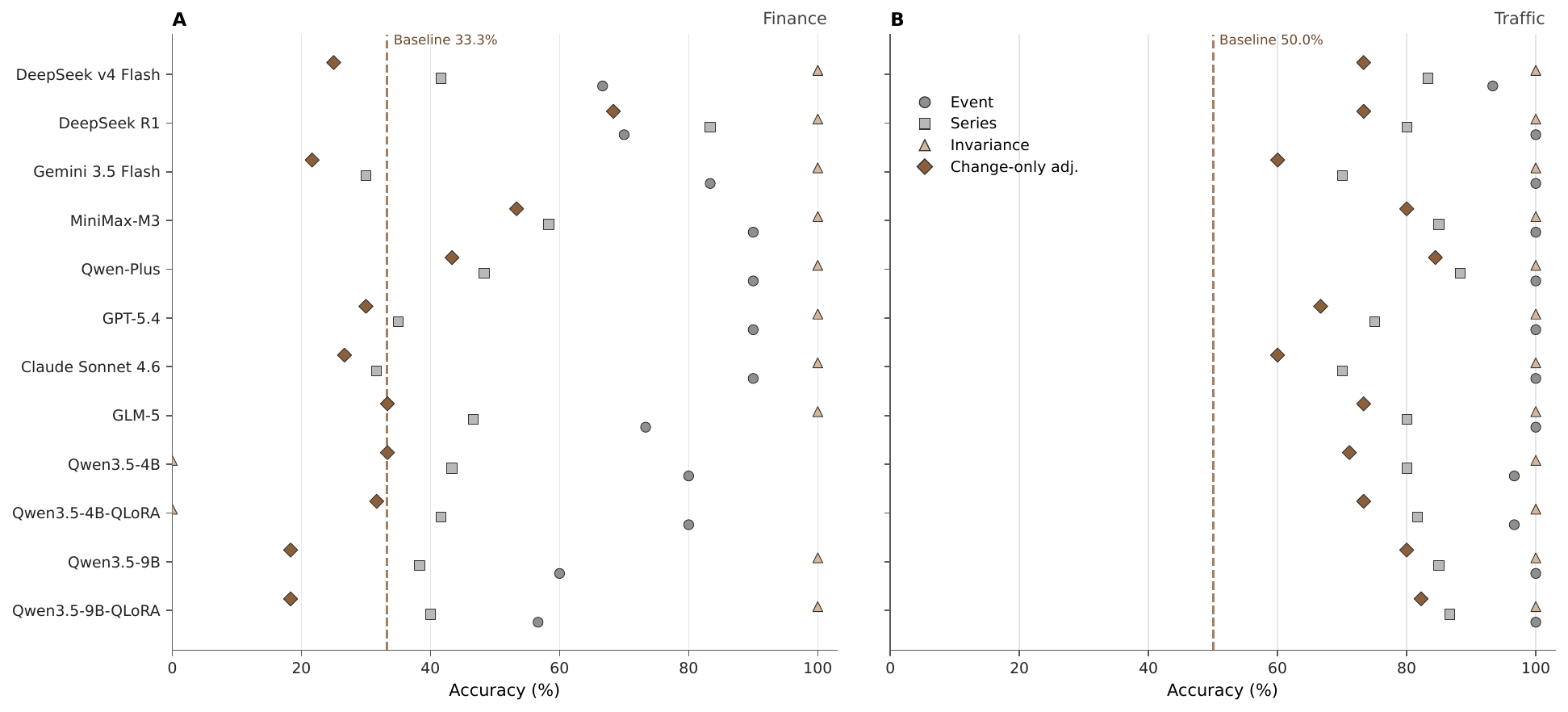}%
}{%
    \fbox{\parbox[c][0.20\textheight][c]{0.96\textwidth}{%
    \centering
    \textbf{Relation-probe category decomposition}\\[4pt]
    Finance and Traffic grouped dot plots for Event, Series, Invariance, and
    change-only adjusted relation accuracy.
    }}%
}
\caption{Relation-probe category decomposition. Event-side, series-side,
invariance, and change-only adjusted accuracy are reported separately.
Change-only accuracy isolates directional transition recognition for the
recognition--application comparison.}
\label{fig:app-relation-breakdown}
\end{figure*}

\subsection{Evidence Validity}
\label{app:evidence-details}

Evidence validity evaluates whether explanations identify required or allowed
variables while avoiding response-irrelevant factors. The decomposition
distinguishes evidence coverage from evidence selectivity.

\begin{table}[t]
\centering
\caption{Evidence precision, recall, and F1 across Finance and Traffic. Each cell reports Finance / Traffic percentages.}
\begin{tabular}{@{}lccc@{}}
\toprule
\textbf{Model}
& \textbf{Precision}
& \textbf{Recall}
& \textbf{F1} \\
\midrule
DeepSeek v4 Flash
& 84.3 / 64.7
& 57.3 / 49.5
& 67.8 / 48.8 \\

DeepSeek R1
& 77.3 / 60.3
& 51.7 / 63.4
& 61.5 / 56.4 \\

Gemini 3.5 Flash
& 78.8 / 65.0
& 45.9 / 34.9
& 57.6 / 38.6 \\

MiniMax-M3
& 83.2 / 57.3
& 57.8 / 71.9
& 67.8 / 59.0 \\

Qwen-Plus
& 85.8 / 55.5
& 56.9 / 46.0
& 67.7 / 42.9 \\

GPT-5.4
& 85.6 / 59.9
& 61.0 / 71.5
& 70.9 / 62.9 \\

Claude Sonnet 4.6
& 83.0 / 57.0
& 58.0 / 86.1
& 67.8 / 67.5 \\

GLM-5
& 82.5 / 62.3
& 56.8 / 32.7
& 66.9 / 34.4 \\

Qwen3.5-9B
& 85.6 / 53.4
& 60.7 / 41.5
& 70.6 / 37.5 \\

Qwen3.5-9B-QLoRA
& 87.5 / 55.4
& 62.0 / 44.1
& 72.2 / 39.3 \\
\bottomrule
\end{tabular}
\end{table}

The contrast between evidence recall and forbidden-factor citation shows why
coverage alone cannot characterize explanation quality. TimeLitmus therefore
reports evidence selection, coverage, and forbidden-cue reliance separately.

\subsection{Linked Behavioral Support}
\label{app:linked-details}

Linked diagnostics evaluate whether correct prediction behavior persists when
the explanation is connected to the corresponding evidence or hard contrast.
Table~\ref{tab:app-linked-ci} reports Basic-linked and HPC-linked Behavioral
Support with bootstrap intervals.

\begin{table*}[t]
\centering
\caption{Linked Behavioral Support and Basic-minus-HPC gaps with 95\%
bootstrap intervals. Basic and HPC estimates use their respective linked
sets.}
\label{tab:app-linked-ci}
\scriptsize
\setlength{\tabcolsep}{2.4pt}
\resizebox{\textwidth}{!}{%
\begin{tabular}{lcccccc}
\toprule
\textbf{Model}
& \textbf{F Basic}
& \textbf{F HPC}
& \textbf{F Gap}
& \textbf{T Basic}
& \textbf{T HPC}
& \textbf{T Gap} \\
\midrule
DeepSeek v4 Flash & 76.5 [70.5, 82.5] & 36.0 [30.7, 41.3] & +40.5 [+32.5, +48.3] & 60.9 [45.7, 73.9] & 21.9 [9.4, 37.5] & +39.0 [+18.5, +59.0] \\
DeepSeek R1 & 87.0 [82.0, 91.5] & 45.3 [39.7, 51.0] & +41.7 [+34.2, +49.0] & 80.4 [69.6, 91.3] & 46.9 [31.2, 62.6] & +33.6 [+12.6, +53.8] \\
Gemini 3.5 Flash & 78.5 [72.5, 84.0] & 41.0 [35.7, 46.7] & +37.5 [+29.5, +45.2] & 47.8 [32.6, 63.0] & 12.5 [3.1, 25.0] & +35.3 [+17.0, +53.4] \\
MiniMax-M3 & 91.0 [87.0, 95.0] & 39.0 [33.3, 44.3] & +52.0 [+45.2, +58.7] & 80.4 [67.4, 91.3] & 31.2 [15.6, 46.9] & +49.2 [+29.2, +68.2] \\
Qwen-Plus & 50.5 [43.5, 57.5] & 43.0 [37.3, 48.7] & +7.5 [-1.3, +16.3] & 87.0 [76.1, 95.7] & 31.2 [15.6, 46.9] & +55.7 [+36.7, +73.8] \\
GPT-5.4 & 78.0 [72.0, 83.5] & 39.0 [33.7, 44.3] & +39.0 [+31.0, +46.8] & 78.3 [65.2, 89.1] & 12.5 [3.1, 25.0] & +65.8 [+48.6, +81.7] \\
Claude Sonnet 4.6 & 57.5 [50.5, 64.5] & 43.0 [37.3, 48.7] & +14.5 [+5.7, +23.5] & 76.1 [63.0, 87.0] & 46.9 [31.2, 65.6] & +29.2 [+7.7, +50.4] \\
GLM-5 & 88.5 [84.0, 93.0] & 43.3 [37.7, 49.0] & +45.2 [+38.0, +52.2] & 67.4 [54.3, 80.4] & 25.0 [9.4, 40.6] & +42.4 [+21.9, +61.7] \\
Qwen3.5-9B & 45.0 [38.5, 52.0] & 44.0 [38.3, 49.7] & +1.0 [-7.8, +9.8] & 89.1 [80.4, 97.8] & 59.4 [43.8, 75.0] & +29.8 [+10.7, +48.8] \\
Qwen3.5-4B & 48.0 [41.0, 55.0] & 36.0 [30.7, 41.3] & +12.0 [+3.3, +20.7] & 56.5 [41.3, 69.6] & 21.9 [9.4, 37.5] & +34.6 [+13.7, +54.6] \\
\bottomrule
\end{tabular}}
\end{table*}

The consistent reduction from Basic-linked to HPC-linked support shows that
behaviorally supported explanation becomes substantially harder under natural
hard contrasts.

\begin{figure*}[t]
\centering
\IfFileExists{A3-1.pdf}{%
    \includegraphics[width=0.98\textwidth]{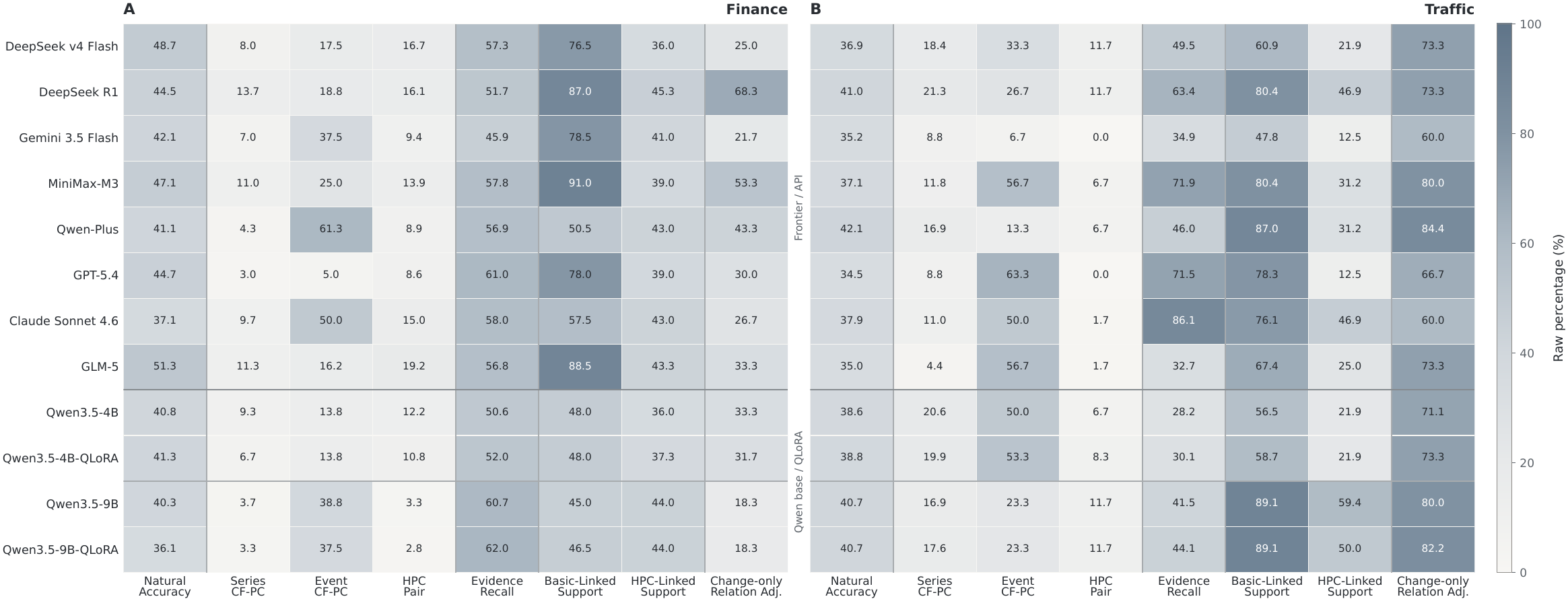}%
}{%
    \fbox{\parbox[c][0.22\textheight][c]{0.96\textwidth}{%
    \centering
    \textbf{Full diagnostic model profiles}\\[4pt]
    Finance and Traffic heatmaps with models as rows and diagnostic metrics as
    columns.
    }}%
}
\caption{Full diagnostic model profiles. The heatmaps preserve the
multidimensional structure of TimeLitmus by showing Natural Accuracy,
series-side and event-side CF-PC, HPC Pair, Evidence Recall, Basic-linked
Support, HPC-linked Support, and change-only Relation Accuracy without
collapsing them into a single score.}
\label{fig:app-diagnostic-heatmap}
\end{figure*}

\subsection{Shortcut-Control Results}
\label{app:shortcut-details}

Partial-modality and cue controls identify performance attainable without the
intended joint use of event and series inputs. Finance reports absolute
event-only and series-only accuracy together with matched hint sensitivity.
Traffic reports matched full-input-minus-control effects. Positive matched
effects indicate degradation after removing a modality; negative effects
indicate that the control performs as well as or better than the full-input
condition.

\begin{table*}[t]
\centering
\caption{Shortcut-control results. Finance Event-only and Series-only columns
report absolute control accuracy. The remaining $\Delta$ columns report
full-input or reference accuracy minus control accuracy in percentage points.}
\label{tab:app-shortcuts}
\scriptsize
\setlength{\tabcolsep}{3pt}
\resizebox{0.8\textwidth}{!}{%
\begin{tabular}{lrrrrr}
\toprule
\textbf{Model}
& \textbf{F Event-only Acc.}
& \textbf{F Series-only Acc.}
& \textbf{F Hint $\Delta$}
& \textbf{T Event-only $\Delta$}
& \textbf{T Series-only $\Delta$} \\
\midrule
DeepSeek v4 Flash & 48.0 & 48.0 & +4.0 & +20.0 & -16.7 \\
DeepSeek R1 & 40.0 & 26.0 & -3.0 & +10.0 & -33.3 \\
Gemini 3.5 Flash & 38.0 & 14.0 & +12.0 & +0.0 & +0.0 \\
MiniMax-M3 & 42.0 & 52.0 & +6.0 & +13.3 & -23.3 \\
Qwen-Plus & 38.0 & 36.0 & +10.0 & +23.3 & -50.0 \\
GPT-5.4 & 48.0 & 50.0 & +1.0 & +0.0 & +0.0 \\
Claude Sonnet 4.6 & 52.0 & 22.0 & -1.0 & +3.3 & -30.0 \\
GLM-5 & 40.0 & 40.0 & +12.0 & +13.3 & -16.7 \\
Qwen3.5-4B & 42.0 & 42.0 & +8.0 & +20.0 & -20.0 \\
Qwen3.5-4B-QLoRA & 42.0 & 42.0 & +8.0 & +20.0 & -20.0 \\
Qwen3.5-9B & 36.0 & 44.0 & +5.0 & -16.7 & -6.7 \\
Qwen3.5-9B-QLoRA & 36.0 & 44.0 & +16.0 & -13.3 & -3.3 \\
\bottomrule
\end{tabular}}
\end{table*}

Traffic series-only controls frequently match or exceed full-input accuracy,
revealing a strong historical-series shortcut that natural accuracy alone
cannot isolate. Finance controls similarly show that nontrivial performance
can be sustained from a single modality.

\begin{figure*}[t]
\centering
\IfFileExists{A4.pdf}{%
    \includegraphics[width=0.98\textwidth]{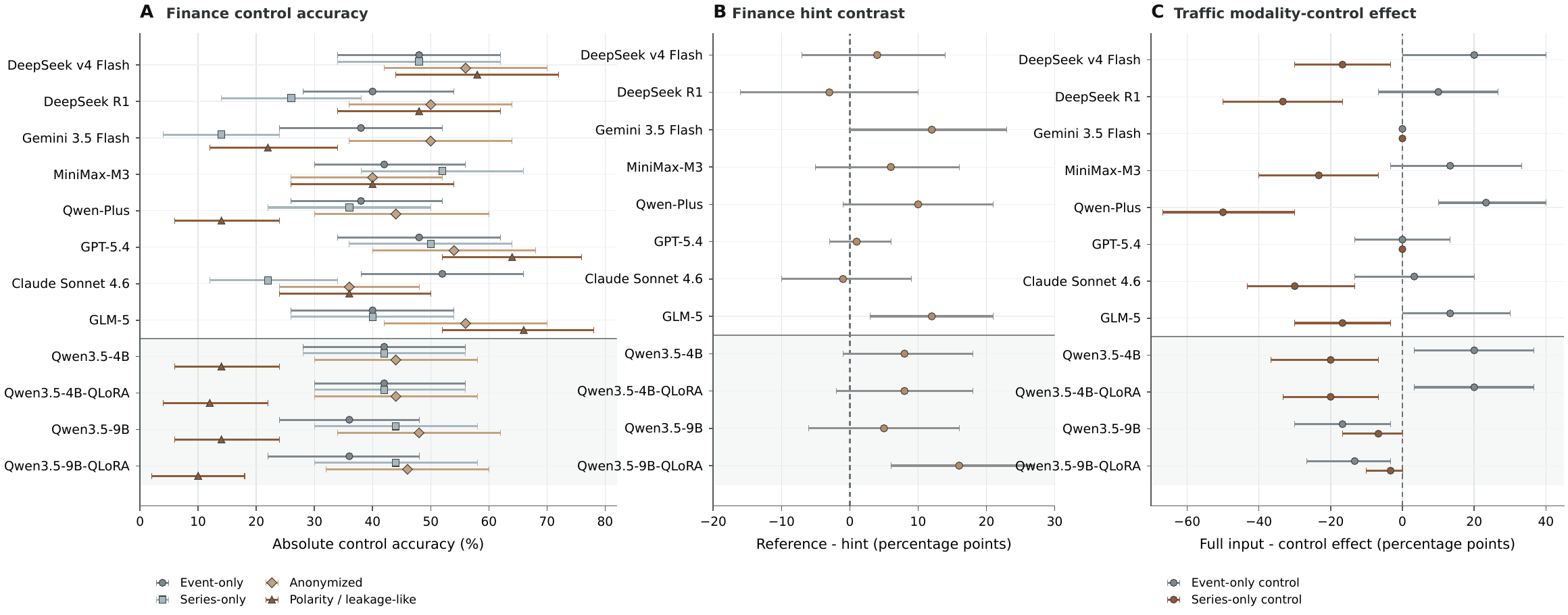}%
}{%
    \fbox{\parbox[c][0.23\textheight][c]{0.96\textwidth}{%
    \centering
    \textbf{Shortcut-control profiles}\\[4pt]
    Finance partial-modality accuracy and hint sensitivity; Traffic matched
    full-input-minus-control effects.
    }}%
}
\caption{Shortcut-control profiles. Finance reports accuracy under
partial-modality and cue-control inputs, while Traffic reports matched
full-input-minus-control effects. Together, the panels identify performance
attainable through unimodal or superficial cues.}
\label{fig:app-shortcut-heatmap}
\end{figure*}

\section{Natural-Only Adaptation Study}
\label{app:adaptation-attribution}

This appendix documents the natural-only adaptation corpus, QLoRA
configuration, checkpoint-selection protocol, and seed-averaged paired effects.
The study evaluates whether supervision on standard event-conditioned
prediction transfers to the controlled, contrastive, and explanation-linked
capabilities measured by TimeLitmus.

\subsection{Adaptation Corpus and Diagnostic Isolation}
\label{app:natural-adaptation-corpus}

The adaptation corpus contains only natural event--series prediction instances
with rule-grounded answers, evidence variables, and concise supporting
rationales. Controlled counterfactual pairs, label-preserving pairs, Hard
Paired Contrasts, linked-faithfulness records, shortcut controls, auxiliary
relation probes, and all TimeLitmus evaluation answers are excluded from
training and checkpoint selection.

We further enforce source-level separation between the adaptation corpus and
the diagnostic benchmark. Finance candidates are removed when they share a
source-news identifier, ticker--date combination, near-duplicate headline,
event window, or price window with the diagnostic release. Traffic candidates
are removed when they share an incident identifier,
freeway--direction--postmile--time combination, near-duplicate incident
description, sensor identifier, or pre-report sensor window with the
diagnostic release.

The resulting training set contains 3,000 instances derived from 2,549 unique
sources, with 451 additional instances reserved for natural-prediction
development. This construction ensures that adaptation uses natural-task
supervision while the controlled relations, hard contrasts, evidence links,
and shortcut annotations evaluated by TimeLitmus remain unseen.

\subsection{QLoRA Configuration}
\label{app:qlora-configuration}

We adapt Qwen3.5-4B and Qwen3.5-9B using the same QLoRA configuration. For each
model scale, we train three independent runs with seeds
$\{42,123,1234\}$. Within each run, checkpoint selection uses held-out
natural-prediction development accuracy. The selected checkpoint from each run
is evaluated on the complete diagnostic suite, and all reported QLoRA results
are averaged across the three selected runs.

Both model scales use 4-bit NF4 quantization with double quantization and BF16
computation. LoRA adapters are applied to the attention and MLP projection
layers.

\begin{table*}[t]
\centering
\caption{QLoRA training and selection configuration.}
\label{tab:qlora-config}
\scriptsize
\setlength{\tabcolsep}{3pt}
\resizebox{\textwidth}{!}{%
\begin{tabular}{lll}
\toprule
\textbf{Configuration}
& \textbf{Qwen3.5-4B-QLoRA}
& \textbf{Qwen3.5-9B-QLoRA} \\
\midrule
Backbone repository
& Qwen/Qwen3.5-4B
& Qwen/Qwen3.5-9B \\
Backbone revision
& 851bf6e806efd8d0a36b00ddf55e13ccb7b8cd0a
& c202236235762e1c871ad0ccb60c8ee5ba337b9a \\
Quantization
& 4-bit NF4, double quantization
& 4-bit NF4, double quantization \\
Compute precision
& BF16
& BF16 \\
LoRA rank / alpha / dropout
& 16 / 32 / 0.05
& 16 / 32 / 0.05 \\
Target modules
& Attention and MLP projections
& Attention and MLP projections \\
Learning rate / scheduler
& $2\times10^{-4}$ / linear
& $2\times10^{-4}$ / linear \\
Warmup
& ratio 0.03
& ratio 0.03 \\
Epochs / maximum optimizer steps
& 2 / 376
& 2 / 376 \\
Micro-batch / accumulation / effective batch
& 1 / 16 / 16
& 1 / 16 / 16 \\
Maximum sequence length
& 4096
& 4096 \\
Optimizer / weight decay
& fused AdamW / 0
& fused AdamW / 0 \\
Gradient checkpointing
& enabled
& enabled \\
Training seeds
& $\{42,123,1234\}$
& $\{42,123,1234\}$ \\
Data seed
& 20260718
& 20260718 \\
Train / development instances
& 3,000 / 451
& 3,000 / 451 \\
Unique training sources
& 2,549
& 2,549 \\
Checkpoint selection
& best development checkpoint per seed
& best development checkpoint per seed \\
Result aggregation
& mean over three selected runs
& mean over three selected runs \\
\bottomrule
\end{tabular}}
\end{table*}

\subsection{Paired Evaluation Protocol}
\label{app:paired-adaptation-results}

Base and QLoRA outputs are aligned on identical evaluation units. Alignment is
performed by record identifier for standalone tasks, pair identifier for
controlled and HPC diagnostics, link identifier for linked diagnostics, and
relation identifier for auxiliary probes. Identifier checks confirm complete
alignment with no missing or duplicate evaluation units.

For each training seed, Base--QLoRA effects are computed on aligned units. The
reported QLoRA score and paired effect are averaged across the three
independently trained runs. Bootstrap resampling preserves evaluation-unit
pairing within each seed and aggregates the paired effects across seeds.

Table~\ref{tab:complete-adaptation-effects} reports the complete
paired QLoRA-minus-Base effects and confidence intervals underlying
the main-text adaptation analysis.

\begin{table*}[t]
\centering
\caption{
Paired QLoRA-minus-Base effects with 95\% paired-bootstrap
intervals. Auxiliary Relation
reports full-probe raw accuracy.
}
\label{tab:complete-adaptation-effects}

\scriptsize
\setlength{\tabcolsep}{2.0pt}
\renewcommand{\arraystretch}{0.96}

\resizebox{0.7\textwidth}{!}{%
\begin{tabular}{@{}llcccc@{}}
\toprule
\multicolumn{6}{c}{\textbf{A: Prediction and controlled behavior}} \\
\cmidrule(lr){1-6}
\textbf{Scale}
& \textbf{Domain}
& $\Delta$ \textbf{Natural}
& $\Delta$ \textbf{Series $\Delta_{\mathrm{CF}}$}
& $\Delta$ \textbf{Event $\Delta_{\mathrm{CF}}$}
& $\Delta$ \textbf{HPC Pair} \\
\midrule
4B & Finance
& +0.5 [-0.5, +1.6]
& -1.8 [-3.56, -0.04]
& +1.0 [-1.6, +3.8]
& -1.4 [-2.8, 0.0] \\
4B & Traffic
& +0.2 [-0.5, +0.9]
& -0.7 [-2.4, +1.1]
& +1.3 [0.0, +4.3]
& +1.7 [0.0, +5.0] \\
9B & Finance
& -4.2 [-7.1, -1.3]
& 0.0 [-0.9, +0.9]
& -2.1 [-5.6, +1.2]
& -0.6 [-2.2, +1.1] \\
9B & Traffic
& 0.0 [-1.7, +1.7]
& +2.1 [-0.1, +4.7]
& 0.0 [0.0, 0.0]
& 0.0 [0.0, 0.0] \\
\midrule
\multicolumn{6}{c}{\textbf{B: Evidence and linked diagnostics}} \\
\cmidrule(lr){1-6}
\textbf{Scale}
& \textbf{Domain}
& $\Delta$ \textbf{Evi. Recall}
& $\Delta$ \textbf{Basic Support}
& $\Delta$ \textbf{HPC Support}
& $\Delta$ \textbf{Aux. Relation} \\
\midrule
4B & Finance
& +1.3 [+0.10, +2.58]
& 0.0 [-2.0, +2.0]
& +1.3 [-1.3, +4.0]
& -0.8 [-3.3, +1.7] \\
4B & Traffic
& +1.9 [-0.8, +3.9]
& +2.2 [-4.3, +8.7]
& 0.0 [0.0, 0.0]
& +0.8 [0.0, +2.5] \\
9B & Finance
& +1.4 [-0.7, +3.6]
& +1.5 [0.0, +3.5]
& 0.0 [0.0, 0.0]
& 0.0 [-2.5, +2.5] \\
9B & Traffic
& +2.6 [0.0, +6.1]
& 0.0 [0.0, 0.0]
& -9.4 [-21.9, 0.0]
& +0.8 [0.0, +2.5] \\
\bottomrule
\end{tabular}%
}
\end{table*}

\subsection{Qwen3.5-4B Adaptation Results}
\label{app:adaptation-4b}

Table~\ref{tab:paired-4b-finance} report the complete 4B adaptation profile across
natural prediction, controlled consistency, hard contrasts, evidence validity,
linked support, and relation recognition.

\begin{table*}[t]
\centering
\caption{Paired Base--QLoRA results for Qwen3.5-4B on Finance. QLoRA scores are
averaged across three independently trained runs with seeds
$\{42,123,1234\}$ after development-set checkpoint selection. Scores,
changes, and confidence intervals are percentage points.}
\label{tab:paired-4b-finance}

\small
\setlength{\tabcolsep}{7pt}
\renewcommand{\arraystretch}{1.05}

\begin{tabular}{@{}lrrrr@{}}
\toprule
\textbf{Metric}
& \textbf{Base}
& \textbf{QLoRA}
& \textbf{$\Delta$}
& \textbf{Paired 95\% CI} \\
\midrule
Natural Accuracy
& 40.8 & 41.3 & +0.5 & $[-0.53,+1.58]$ \\
Series CF-PC
& 9.3 & 6.7 & -2.7 & $[-5.33,+0.00]$ \\
Series $\Delta_{\mathrm{CF}}$
& -10.0 & -11.8 & -1.8 & $[-3.56,-0.04]$ \\
Event CF-PC
& 13.8 & 13.8 & +0.0 & $[-3.75,+3.75]$ \\
Event $\Delta_{\mathrm{CF}}$
& -2.0 & -1.0 & +1.0 & $[-1.56,+3.75]$ \\
HPC Pair Correctness
& 12.2 & 10.8 & -1.4 & $[-2.78,+0.00]$ \\
Evidence Precision
& 78.6 & 79.6 & +1.0 & $[-0.46,+2.46]$ \\
Evidence Recall
& 50.6 & 52.0 & +1.3 & $[+0.10,+2.58]$ \\
Evidence F1
& 60.8 & 62.2 & +1.3 & $[+0.04,+2.63]$ \\
Basic-Linked Behavioral Support
& 48.0 & 48.0 & +0.0 & $[-2.00,+2.00]$ \\
HPC-Linked Behavioral Support
& 36.0 & 37.3 & +1.3 & $[-1.33,+4.00]$ \\
Auxiliary Relation Accuracy
& 66.7 & 65.8 & -0.8 & $[-3.33,+1.67]$ \\
\bottomrule
\end{tabular}
\end{table*}

The 4B Finance model improves required-evidence recall and Evidence F1 after
natural-only adaptation. The controlled and hard-pair metrics remain largely
unchanged, showing that evidence-selection gains and cross-modal behavioral
consistency are distinct adaptation outcomes.

\subsection{Qwen3.5-9B Adaptation Results}
\label{app:adaptation-9b}

Tables~\ref{tab:paired-9b-finance} and
\ref{tab:paired-9b-traffic} report the corresponding 9B results under the same
training, checkpoint-selection, and paired-evaluation protocol.

\begin{table*}[t]
\centering
\caption{Paired Base--QLoRA results for Qwen3.5-9B on Finance. QLoRA scores are
averaged across three independently trained runs with seeds
$\{42,123,1234\}$ after development-set checkpoint selection. Scores,
changes, and confidence intervals are percentage points.}
\label{tab:paired-9b-finance}

\small
\setlength{\tabcolsep}{7pt}
\renewcommand{\arraystretch}{1.05}

\begin{tabular}{@{}lrrrr@{}}
\toprule
\textbf{Metric}
& \textbf{Base}
& \textbf{QLoRA}
& \textbf{$\Delta$}
& \textbf{Paired 95\% CI} \\
\midrule
Natural Accuracy
& 40.3 & 36.1 & -4.2 & $[-7.11,-1.32]$ \\
Series CF-PC
& 3.7 & 3.3 & -0.3 & $[-1.00,+0.00]$ \\
Series $\Delta_{\mathrm{CF}}$
& -15.9 & -16.0 & -0.0 & $[-0.92,+0.95]$ \\
Event CF-PC
& 38.8 & 37.5 & -1.3 & $[-6.25,+2.50]$ \\
Event $\Delta_{\mathrm{CF}}$
& -3.5 & -5.6 & -2.1 & $[-5.59,+1.25]$ \\
HPC Pair Correctness
& 3.3 & 2.8 & -0.6 & $[-2.22,+1.11]$ \\
Evidence Precision
& 85.6 & 87.5 & +1.9 & $[-0.45,+4.38]$ \\
Evidence Recall
& 60.7 & 62.0 & +1.4 & $[-0.73,+3.58]$ \\
Evidence F1
& 70.6 & 72.2 & +1.6 & $[-0.58,+3.90]$ \\
Basic-Linked Behavioral Support
& 45.0 & 46.5 & +1.5 & $[+0.00,+3.50]$ \\
HPC-Linked Behavioral Support
& 44.0 & 44.0 & +0.0 & $[+0.00,+0.00]$ \\
Auxiliary Relation Accuracy
& 59.2 & 59.2 & +0.0 & $[-2.50,+2.50]$ \\
\bottomrule
\end{tabular}
\end{table*}

\begin{table*}[t]
\centering
\caption{Paired Base--QLoRA results for Qwen3.5-9B on Traffic. QLoRA scores are
averaged across three independently trained runs with seeds
$\{42,123,1234\}$ after development-set checkpoint selection. Scores,
changes, and confidence intervals are percentage points.}
\label{tab:paired-9b-traffic}

\small
\setlength{\tabcolsep}{7pt}
\renewcommand{\arraystretch}{1.05}

\begin{tabular}{@{}lrrrr@{}}
\toprule
\textbf{Metric}
& \textbf{Base}
& \textbf{QLoRA}
& \textbf{$\Delta$}
& \textbf{Paired 95\% CI} \\
\midrule
Natural Accuracy
& 40.7 & 40.7 & +0.0 & $[-1.67,+1.67]$ \\
Series CF-PC
& 16.9 & 17.6 & +0.7 & $[-1.47,+3.68]$ \\
Series $\Delta_{\mathrm{CF}}$
& -4.5 & -2.3 & +2.1 & $[-0.12,+4.65]$ \\
Event CF-PC
& 23.3 & 23.3 & +0.0 & $[+0.00,+0.00]$ \\
Event $\Delta_{\mathrm{CF}}$
& -3.3 & -3.3 & +0.0 & $[-0.00,-0.00]$ \\
HPC Pair Correctness
& 11.7 & 11.7 & +0.0 & $[+0.00,+0.00]$ \\
Evidence Precision
& 53.4 & 55.4 & +2.0 & $[-0.78,+4.82]$ \\
Evidence Recall
& 41.5 & 44.1 & +2.6 & $[-0.00,+6.08]$ \\
Evidence F1
& 37.5 & 39.3 & +1.9 & $[-0.48,+4.45]$ \\
Basic-Linked Behavioral Support
& 89.1 & 89.1 & +0.0 & $[+0.00,+0.00]$ \\
HPC-Linked Behavioral Support
& 59.4 & 50.0 & -9.4 & $[-21.88,+0.00]$ \\
Auxiliary Relation Accuracy
& 92.5 & 93.3 & +0.8 & $[+0.00,+2.50]$ \\
\bottomrule
\end{tabular}
\end{table*}

Across both model scales, natural-only adaptation produces selective changes
in natural prediction and evidence selection but does not yield a consistent
improvement in controlled consistency, HPC correctness, or linked behavioral
support. TimeLitmus therefore distinguishes task-compatible adaptation from
transfer to the cross-modal capabilities required by its diagnostics.

\subsection{Statistical Analysis}
\label{app:adaptation-statistics}

We use 2,000 bootstrap iterations with bootstrap seed 42. Natural prediction
and evidence metrics resample records; CF-PC and
$\Delta_{\mathrm{CF}}$ resample complete controlled pairs; HPC resamples
complete hard pairs; linked metrics resample complete links; and relation
probes resample relation records. For $\Delta_{\mathrm{CF}}$, the
endpoint-independence baseline is recomputed within every bootstrap draw.

Within each training seed, Base and QLoRA outputs are resampled as aligned
pairs. Paired effects are computed per seed and averaged across the three
independently trained runs. Finance and Traffic are analyzed separately.

\subsection{Output-Logit Sensitivity Analysis}
\label{app:logit-sensitivity}

We complement the prediction- and explanation-level diagnostics in
TimeLitmus with an output-logit sensitivity analysis for the
open-weight Qwen3.5-4B and Qwen3.5-9B models.
While the main controlled diagnostics evaluate whether models produce
relation-consistent predictions, this analysis measures how strongly
different input components support the model's confidence in the gold
answer. It therefore provides an additional diagnostic layer between
evidence citation and pair-level behavioral application.

\paragraph{Evaluation subset.}
For each model size, we use a shared subset of 720 instances, comprising
360 Finance and 360 Traffic records.
The Finance subset contains 120 Natural Prediction, 120 controlled, and
120 evidence-attribution instances.
The Traffic subset contains 135 Natural Prediction, 161 controlled, and
64 evidence-attribution instances.
The same instance set is used for Base and QLoRA evaluation.

For every instance, we construct the following input variants:
the complete input, event-masked input, series-masked input,
recent-window-masked input, required-evidence-masked input, and
forbidden-cue-masked input.
Mask-specific results are computed only for instances to which the
corresponding intervention applies.

\paragraph{Forced-choice scoring.}
We score the next-token probabilities assigned to the answer choices:
A--D for Finance and A--C for Traffic.
Let $\mathcal{Y}$ denote the corresponding answer-choice set,
$y_i^\ast$ the gold choice for instance $i$, $x_i$ the complete input,
and $x_{i,\setminus m}$ the input after masking component $m$.
We define the input importance of component $m$ as
\begin{equation}
I_{i,m}
=
\log P_{\mathcal{Y}}
\left(
y_i^\ast \mid x_i
\right)
-
\log P_{\mathcal{Y}}
\left(
y_i^\ast \mid x_{i,\setminus m}
\right),
\label{eq:logprob-importance}
\end{equation}
where $P_{\mathcal{Y}}$ is normalized over the candidate answer choices.
A positive importance value indicates that the removed input component
supports the gold-label prediction.

We evaluate three independently trained QLoRA models with seeds
$\{42,123,1234\}$.
For each instance and masking condition, QLoRA results are first
computed separately for each seed and then averaged across the three
runs.
QLoRA-minus-Base effects are estimated on the shared instances.
We report paired 95\% confidence intervals using 2,000 bootstrap
resamples.

\paragraph{Results.}
Table~\ref{tab:output-logit-sensitivity} reports the attribution effects
most directly associated with event--series evidence use.
The Qwen3.5-4B results exhibit a particularly structured pattern.
In Finance, QLoRA increases sensitivity to both the textual event and
the historical series, while also increasing the importance of inputs
associated with required evidence.
In Traffic, the principal shift occurs on the series side, including
greater sensitivity to the recent temporal window.
These patterns correspond to the different diagnostic structures of
the two domains: Finance requires the joint interpretation of event
semantics and pre-event market state, whereas Traffic relies strongly
on the recent evolution of the observed series.

\begin{table*}[t]
\centering
\caption{
Selected output-logit sensitivity results for Base and QLoRA models.
QLoRA values are averaged across three independently trained runs.
Accuracy changes are reported in percentage points; importance changes
are differences in forced-choice gold-label log-probability.
Brackets report paired 95\% bootstrap confidence intervals.
}
\label{tab:output-logit-sensitivity}

\scriptsize
\setlength{\tabcolsep}{4.5pt}
\renewcommand{\arraystretch}{1.10}

\begin{tabular*}{0.8\textwidth}{
@{\extracolsep{\fill}}lllrll@{}
}
\toprule
\textbf{Model}
& \textbf{Domain / Subset}
& \textbf{Metric}
& \textbf{Base}
& \textbf{QLoRA}
& \textbf{$\Delta$ [95\% CI]} \\
\midrule

\multirow{7}{*}{Qwen3.5-4B}
& \multirow{4}{*}{Finance / All}
& Forced-choice Accuracy
& 30.6
& 38.1
& $\mathbf{+7.5}$ $[+3.1,+11.9]$ pp \\

&
& Event Importance
& 0.068
& 0.287
& $\mathbf{+0.220}$ $[+0.173,+0.273]$ \\

&
& Series Importance
& 0.064
& 0.134
& $\mathbf{+0.071}$ $[+0.032,+0.113]$ \\

&
& Required-Evidence Importance
& 0.140
& 0.323
& $\mathbf{+0.183}$ $[+0.058,+0.310]$ \\
\cmidrule(lr){2-6}

& \multirow{3}{*}{Traffic / All}
& Forced-choice Accuracy
& 50.8
& 56.7
& $\mathbf{+5.8}$ $[+2.2,+9.7]$ pp \\

&
& Series Importance
& 0.002
& 0.089
& $\mathbf{+0.088}$ $[+0.052,+0.123]$ \\

&
& Recent-Window Importance
& $-0.033$
& $-0.005$
& $\mathbf{+0.028}$ $[+0.010,+0.045]$ \\

\midrule

\multirow{6}{*}{Qwen3.5-9B}
& \multirow{2}{*}{Finance / All}
& Forced-choice Accuracy
& 34.7
& 44.7
& $\mathbf{+10.0}$ $[+1.9,+18.1]$ pp \\

&
& Required-Evidence Importance
& 0.118
& 0.375
& $\mathbf{+0.257}$ $[+0.017,+0.497]$ \\
\cmidrule(lr){2-6}

& Finance / Controlled
& Series Importance
& 0.059
& 0.387
& $\mathbf{+0.329}$ $[+0.170,+0.469]$ \\
\cmidrule(lr){2-6}

& \multirow{3}{*}{Traffic / Controlled}
& Series Importance
& 0.010
& 0.100
& $\mathbf{+0.090}$ $[+0.004,+0.180]$ \\

&
& Recent-Window Importance
& $-0.172$
& $-0.007$
& $\mathbf{+0.166}$ $[+0.101,+0.239]$ \\

&
& Required-Evidence Importance
& $-0.038$
& 0.207
& $\mathbf{+0.245}$ $[+0.135,+0.356]$ \\

\bottomrule
\end{tabular*}
\end{table*}

The larger model exhibits more localized attribution changes.
For Finance, Qwen3.5-9B shows increased required-evidence importance
over the complete evaluation subset and increased series importance on
controlled instances.
For Traffic controlled instances, QLoRA strengthens sensitivity to the
historical series, the recent temporal window, and inputs associated
with required evidence.
The stratified results demonstrate that output-logit attribution can
localize adaptation effects by modality, domain, and diagnostic family,
rather than representing them with a single aggregate score.

\paragraph{Complementarity with TimeLitmus diagnostics.}
The output-logit analysis extends the diagnostic hierarchy of
TimeLitmus along three complementary levels.
Evidence metrics evaluate which factors models identify or cite;
log-probability attribution measures which input components support
gold-label confidence; and CF-PC, HPC Pair Correctness, and linked
behavioral support evaluate whether the corresponding relations are
applied consistently in prediction.
The resulting separation shows that evidence recognition,
confidence-level input sensitivity, and relation-consistent application
constitute distinct diagnostic dimensions.

Together, these results extend TimeLitmus from evaluating what evidence
models mention and how they behave under controlled pairs to examining
which event and time-series inputs support their output confidence.
Complete item-level attribution scores and per-stratum summaries are
included in the released evaluation suite.

\end{document}